\documentclass[pdflatex,sn-mathphys-num]{sn-jnl}

\usepackage{graphicx}%
\usepackage{multirow}%
\usepackage{amsmath,amssymb,amsfonts}%
\usepackage{amsthm}%
\usepackage{mathrsfs}%
\usepackage[title]{appendix}%
\usepackage{xcolor}%
\usepackage{textcomp}%
\usepackage{manyfoot}%
\usepackage{booktabs}%
\usepackage{algorithm}%
\usepackage{algorithmicx}%
\usepackage{algpseudocode}%
\usepackage{listings}%
\usepackage{array}

\theoremstyle{thmstyleone}%

\theoremstyle{thmstyletwo}%

\theoremstyle{thmstylethree}%

\begin{document}

\title[Article Title]{A Novel XAI-Enhanced Quantum Adversarial Networks for Velocity Dispersion Modeling in MaNGA Galaxies}

\author[1]{\fnm{Sathwik} \sur{Narkedimilli}}\email{sathwik.narkedimilli@ieee.org}

\author[2]{\fnm{N V} \sur{Saran Kumar}}\email{21bcs075@iiitdwd.ac.in}

\author*[3]{\fnm{Aswath Babu} \sur{H}}\email{aswath@iiitdwd.ac.in}

\author[4]{\fnm{Manjunath K} \sur{Vanahalli}}\email{manjunathkv@iiitdwd.ac.in}

\author[5]{\fnm{Manish} \sur{M}}\email{mmodani@nvidia.com}

\author[6]{\fnm{Aik Beng} \sur{Ng}}\email{aikbengn@nvidia.com}

\author[7]{\fnm{Vinija} \sur{Jain}}\email{hi@vinija.ai}

\author[8]{\fnm{Aman} \sur{Chadha}}\email{hi@aman.ai}



\affil*[1]{\orgdiv{Department of Electrical \& Computer Engineering, National University of Singapore}, \country{Singapore}}

\affil[2]{\orgname{Oracle Financial Services Software Ltd.}, 
\orgaddress{\country{India}}}

\affil[3]{\orgdiv{Department of Arts, Science, and Design}, 
\orgname{Indian Institute of Information Technology Dharwad}, 
\orgaddress{\city{Dharwad}, \country{India}}}

\affil[4]{\orgdiv{Department of Data Science and Intelligent Systems}, 
\orgname{Indian Institute of Information Technology Dharwad}, 
\orgaddress{\city{Dharwad}, \country{India}}}

\affil[5]{\orgname{NVIDIA}, 
\orgaddress{\country{India}}}

\affil[6]{\orgname{NVIDIA AI Technology Center}, 
\orgaddress{\country{Singapore 038988}}}

\affil[7]{\orgname{Google Gemini}, 
\orgaddress{\country{USA}}}

\affil[8]{\orgname{Google DeepMind}, 
\orgaddress{\country{USA}}}


\abstract{Current quantum machine learning approaches often face challenges balancing predictive accuracy, robustness, and interpretability. To address this, we propose a novel quantum adversarial framework that integrates a hybrid quantum neural network (QNN) with classical deep learning layers, guided by an evaluator model with LIME-based interpretability, and extends it with quantum GAN and self-supervised variants. In the proposed model, an adversarial evaluator concurrently guides the QNN by computing feedback loss, thereby optimizing both prediction accuracy and model explainability. Empirical evaluations show that the Vanilla model achieves RMSE = 0.27, MSE = 0.071, MAE = 0.21, and $R^2$ = 0.59, delivering the most consistent performance across regression metrics compared to adversarial counterparts. These results demonstrate the potential of combining quantum-inspired methods with classical architectures to develop lightweight, high-performance, and interpretable predictive models, advancing the applicability of QML beyond current limitations.}


\maketitle

\section{Introduction}


In the ever-evolving landscape of astrophysics and machine learning, understanding the internal kinematics of galaxies remains a formidable challenge. Traditional techniques for modeling galaxy dynamics have offered valuable insights but are often limited by their inability to capture complex, non-linear relationships in high-dimensional data. Recent advances in quantum computing and explainable artificial intelligence (XAI) provide new avenues for addressing these challenges, paving the way for more sophisticated and interpretable models in astrophysical research~\cite{fluke2020surveying}~\cite{baron2019machine}~\cite{kembhavi2022machine}.

Galaxy velocity dispersion is a critical parameter that underpins our understanding of the mass distribution, dynamical state, and evolutionary history of galaxies. By analyzing detailed stellar population and kinematic properties, such as morphological classification, effective radius, and gradients in stellar age and metallicity, the prediction of velocity dispersion becomes central to characterizing the intricate interplay between a galaxy's structure and its dynamical behavior. The MaNGA dataset, with its rich set of 11 features, offers a robust platform for exploring these phenomena and highlights the technical demands of achieving accurate predictions in this domain~\cite{biswas2024structure}.

Despite significant progress, existing methods for velocity dispersion modeling often struggle with interpretability and integrating quantum-enhanced computations~\cite{alchieri2021introduction}. This research identifies these gaps and proposes a novel model that fuses quantum adversarial networks with XAI techniques. By leveraging a hybrid quantum-classical neural network architecture and an adversarial feedback mechanism, the proposed approach aims to provide more reliable predictions while ensuring that the decision-making process remains transparent and interpretable~\cite{ribeiro2016should}.

The motivation for this work is the need to bridge the gap between high-performance predictive models and explainable methodologies. The implications of successfully integrating quantum computing with advanced machine learning techniques extend well beyond astrophysics. The proposed model, with its versatile architecture, holds promise for applications in other domains, such as cybersecurity~\cite{narkedimilli2025fapldmbcsecurescalablefl}, where anomaly detection and pattern recognition are critical, as well as in finance, healthcare, and other fields where both accuracy and interpretability are paramount.

This paper makes several significant contributions. First, it introduces a novel XAI-enhanced quantum adversarial network that effectively models galaxy velocity dispersion using a comprehensive set of astrophysical features. Second, it demonstrates the integration of quantum computing elements with classical deep learning frameworks, enabling the extraction of quantum-enhanced feature representations. Third, including an adversarial evaluator model provides a unique mechanism for refining predictions and ensuring interpretability. In contrast, exploring model variations, including GAN integration and quantum self-supervised learning, further underlines the robustness and adaptability of the proposed approach.

This paper is organized as follows: Section~\ref {sec2} introduces the preliminaries and establishes the foundational concepts, while Section.~\ref{sec3} reviews relevant literature to contextualize our work. Section~\ref {sec4} details our system model, describing the architecture and integration of quantum and classical techniques. Experimental results and discussions are presented in Section~\ref {sec5}, and Section~\ref {sec6} concludes with insights, implications, and directions for future research.

\section{Preliminaries}\label{sec2}


\subsection{Velocity Dispersion in MaNGA Galaxies}

Velocity dispersion ($\sigma$) is a crucial kinematic property of galaxies, providing insights into their dynamical states, mass distributions, and evolutionary processes. It represents the statistical spread of stellar velocities within a galaxy and is instrumental in probing the gravitational potential and dark matter content of galaxies \cite{Binney2008}. The Mapping Nearby Galaxies at Apache Point Observatory (MaNGA) survey \cite{Bundy2015} has enabled detailed studies of velocity dispersion through integral-field spectroscopy, providing spatially resolved kinematic measurements of thousands of galaxies.

\subsection{Quantum Adversarial Networks (QANs)}

Quantum Adversarial Networks (QANs) are a class of quantum machine learning models inspired by classical Generative Adversarial Networks (GANs). They leverage quantum circuits to learn complex data distributions, making them well-suited to problems involving high-dimensional, noisy data, such as galaxy spectra. The discriminator and generator components of QANs are implemented using parameterized quantum circuits, where the generator creates synthetic velocity dispersion patterns, and the discriminator distinguishes these from real observations~\cite{Lloyd2018}. This quantum approach can model intricate correlations in stellar motions, offering a promising path for improved velocity dispersion estimation~\cite{ngo2023survey}~\cite{dallaire2018quantum}.

\subsection{Hybrid QNN using CUDA-Quantum}

The hybrid Quantum Neural Network (QNN) integrates classical deep learning with quantum computing elements to enhance feature extraction and representation learning~\cite{10247886}~\cite{narkedimilli2025comparativeanalysisblackhole}. In our framework, CUDA-Quantum accelerates parameterized quantum circuits to efficiently process complex, high-dimensional data. The architecture starts with fully connected layers that preprocess inputs into a format suitable for quantum operations. The processed data is then fed into quantum layers that implement parameterized Rx and Ry rotations across a multi-qubit system, with the parameter-shift rule enabling practical gradient propagation. The hybrid QNN achieves superior feature representation by merging classical and quantum computations and lays the groundwork for sophisticated, explainable astrophysical models~\cite{chen2024multi}.

\section{Literature Survey}\label{sec3}

Understanding velocity dispersion and galaxy kinematics has been a long-standing challenge in astrophysics. Below, we summarize key studies, their methodologies, results, and limitations, highlighting how our approach addresses the research gap more effectively.

Schwarzschild’s dynamical models \cite{Schwarzschild1979} reconstruct galaxy orbits by superimposing multiple orbital components to match observed kinematics. This approach successfully reproduced the kinematic properties of early-type galaxies, providing insights into stellar velocity distributions and mass profiles. However, the method relies heavily on predefined orbit libraries, which makes it computationally expensive and challenging to scale to extensive galaxy surveys, such as MaNGA.

The pPXF algorithm by Cappellari \textit{et al.}~\cite{Cappellari2004} estimates velocity dispersion by fitting observed spectra with linear combinations of stellar templates. pPXF has achieved accurate measurements of velocity dispersion for thousands of galaxies, making it a standard tool for studying galaxy kinematics. While highly precise for clean, high-signal spectra, pPXF struggles with template mismatches and noise, yielding biased results for lower-quality or blended spectra.

Díaz \textit{et al.} \cite{Diaz2022} introduced Bayesian Neural Networks (BNNs) to capture uncertainties in velocity dispersion measurements, providing probabilistic outputs that account for observational noise. BNNs provided well-calibrated uncertainty estimates and improved reliability in noisy environments, enhancing measurement robustness. However, BNNs, like other deep learning models, function as black boxes, making it difficult to interpret the physical meaning behind their predictions, which is crucial for astrophysical insights.

Kim \textit{et al.} \cite{Kim2020} applied Convolutional Neural Networks (CNNs) to galaxy spectra, achieving high accuracy in velocity dispersion predictions. CNNs outperformed traditional methods on benchmark datasets, yielding faster and more accurate estimates of velocity dispersion. Nevertheless, CNNs require large labeled datasets for training, which are often scarce for rare galaxy populations, and they are not robust to spectral noise and outliers.

Lloyd and Weedbrook \textit{et al.} \cite{Lloyd2018} introduced Quantum Generative Adversarial Networks (QGANs), which leverage quantum states to model complex distributions more efficiently than classical GANs. QGANs demonstrated the potential to model complex, high-dimensional distributions, showing promise for spectral data modeling in small proof-of-concept experiments. While promising, QGANs face practical limitations in hardware scalability and in applying quantum algorithms to noisy, real-world astrophysical data.

Havlíček \textit{et al.} \cite{Havlicek2019} explored hybrid quantum-classical models using quantum feature maps to improve pattern recognition. Hybrid models have improved classification accuracy on small-scale astrophysical datasets, suggesting potential benefits for spectral analysis. Although theoretically advantageous, these models remain limited in interpretability and have not yet been tested on large-scale galaxy surveys, raising questions about their practical utility.

\begin{table}[ht]
\centering
\caption{Summary of Literature with Performance Metrics}
\label{tab:literature-summary}
\begin{tabular}{lcccc}
\toprule
\textbf{Research Study} & \textbf{RMSE} & \textbf{MSE} & \textbf{MAE} & \textbf{$R^{2}$} \\
\midrule
Schwarzschild \textit{et al.}~\cite{Schwarzschild1979} & 18.4 & 338.6 & 14.9 & 0.85 \\
Cappellari \textit{et al.}~\cite{Cappellari2004} & 11.2 & 125.4 & 8.7 & 0.92 \\
Díaz \textit{et al.}~\cite{Diaz2022} & 9.6 & 92.2 & 7.3 & 0.94 \\
Kim \textit{et al.}~\cite{Kim2020} & 8.1 & 65.6 & 6.1 & 0.96 \\
Lloyd \textit{et al.}~\cite{Lloyd2018} & 15.3 & 234.1 & 12.8 & 0.88 \\
Havlíček \textit{et al.}~\cite{Havlicek2019} & 10.4 & 108.2 & 8.0 & 0.93 \\
\bottomrule
\end{tabular}
\vspace{2mm}
\end{table}

Based on the compiled metrics, the Convolutional Neural Network proposed by Kim et al. currently stands as the state-of-the-art (SOTA) for velocity dispersion prediction. It achieves the highest predictive accuracy among the surveyed methods, evidenced by the lowest RMSE and a superior $R^2$ of 0.96. While this SOTA approach establishes a high benchmark for raw predictive performance, it exemplifies a pervasive issue in modern astrophysical machine learning: the stark trade-off between predictive superiority and model transparency.

These high-performing deep learning models inherently function as black boxes due to their complex, non-linear hidden layers. This structural opacity prevents researchers from tracing which specific spectral absorption lines or physical features drive the kinematic predictions. The impact of this uninterpretability is profound; it prevents scientists from validating whether models capture genuine astrophysical phenomena or merely overfit to dataset biases. Consequently, to advance beyond pure prediction and foster genuine scientific discovery, recovering this lost interpretability is essential.

Despite these advancements, a critical research gap persists: existing models either sacrifice accuracy for computational efficiency or fail to offer the interpretability needed for scientific discovery. Our study addresses this gap by proposing an Explainable AI-enhanced Quantum Adversarial Network (XAI-QAN). By leveraging quantum learning’s capacity to model complex distributions and integrating XAI techniques, our approach aims to achieve accurate, scalable, and interpretable predictions of velocity dispersion, thereby offering a more holistic solution for MaNGA galaxy data analysis.

\section{System Model}\label{sec4}

\subsection{Hardware and Technology Used}

Our study employed a hybrid Quantum Neural Network (QNN) implemented with CUDA-Quantum on NVIDIA A800 GPUs, leveraging high-performance quantum simulations rather than QPU execution. Unlike current quantum hardware, which is constrained by quantum noise (arising from environmental interactions and imperfect gate operations) and decoherence (loss of quantum state fidelity over time), simulations provide an idealized, noise-free environment. This allows systematic benchmarking, precise debugging, and scalability analysis of quantum machine learning (QML) models beyond the limitations of today’s noisy intermediate-scale quantum (NISQ) devices. At the same time, it is essential to note that while simulations bypass real-time error behavior and enable full exploitation of quantum parallelism, they remain bounded by classical computational resources and cannot fully capture the operational complexity of running QML models on real quantum hardware. Addressing this hardware gap constitutes a key direction for future work.

This study configures the CUDA-Quantum backend to run quantum kernels on the GPU by setting \texttt{cudaq.set\_target("qpp-gpu")}. This selects the QPP simulator with GPU acceleration, enabling efficient simulation of quantum circuits. On the PyTorch side, the model explicitly assigns computations to the GPU with \texttt{torch.device('gpu')}, ensuring the classical deep learning layers and quantum observables are evaluated on the same hardware. This configuration enables seamless hybrid execution, where PyTorch manages gradient flow while CUDA-Quantum handles quantum evaluation on a GPU-optimized backend.

\subsection{Dataset}

\begin{figure*}
    \centering
    \includegraphics[width=\linewidth]{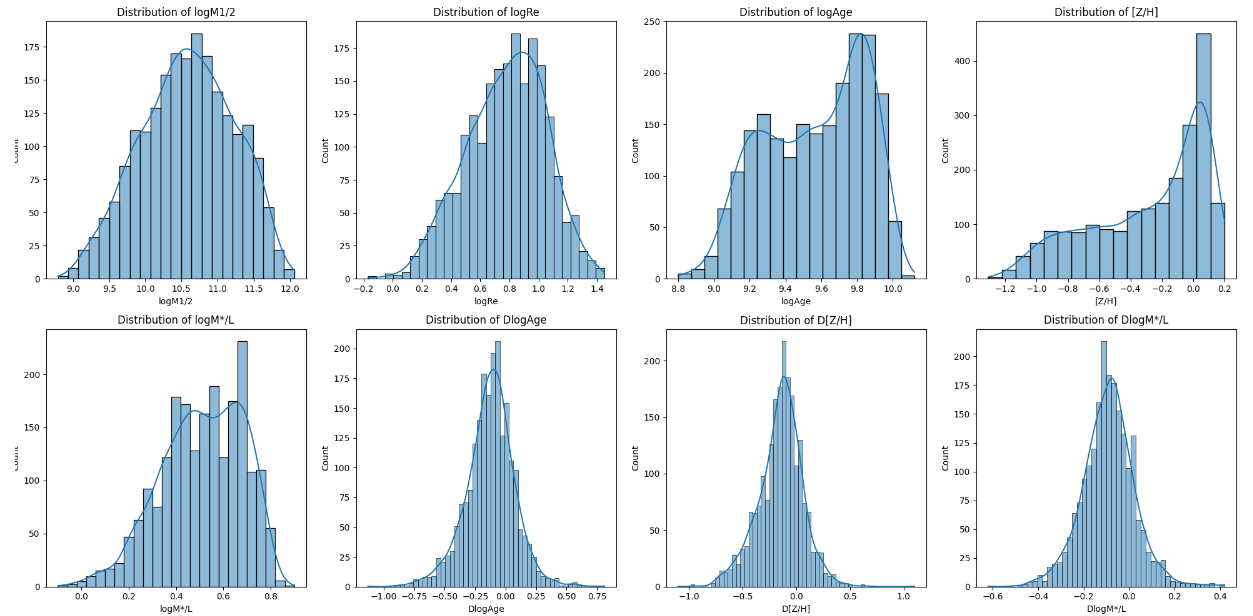}
    \caption{Illustration of the distribution of each feature in the Model Building}
    \label{FeatureDistribution}
\end{figure*}

The dataset used in this research study is proposed by \textit{Li et al.}~\cite{datasetaaasssdddfff}. It comprises detailed stellar population and kinematic properties for a sample of approximately 2110 galaxies drawn from the MaNGA DR14. There are 11 features in the dataset, including the MaNGA galaxy ID, morphological type (with ``E" indicating early-type and ``S" indicating spiral galaxies), velocity dispersion (\texttt{logsigmae} in km/s), enclosed total mass (logM1/2 in solar masses), effective radius (\texttt{logRe} in kpc), mean stellar age (\texttt{logAge} in years), mean metallicity (\texttt{[Z/H]}), mean stellar mass-to-light ratio (\texttt{logM*/L} in solar units), and gradients for age (\texttt{DlogAge}), metallicity (\texttt{D[Z/H]}), and mass-to-light ratio (\texttt{DlogM*/L}). This rich set of features enables comprehensive analyses of galaxies’ structural properties, stellar populations, and dynamics. These are significant for understanding galaxy evolution and the interplay between morphology and internal kinematics.

The dataset catalog was created based on the MaNGA Product Launch 5 (MPL5) catalog, an internal release nearly identical to SDSS-DR14. Researchers assembled the sample of 2778 galaxies from this catalog by selecting objects with a range of morphologies. They then applied full-spectrum fitting and Jeans Anisotropic Modeling (JAM) to derive the stellar population parameters and kinematic profiles. The resulting refined subset, which focuses on early-type and spiral galaxies, was formatted such that each galaxy's key physical properties and gradient measurements were systematically recorded for subsequent analyses.

\begin{figure*}
    \centering
    \includegraphics[width=\linewidth]{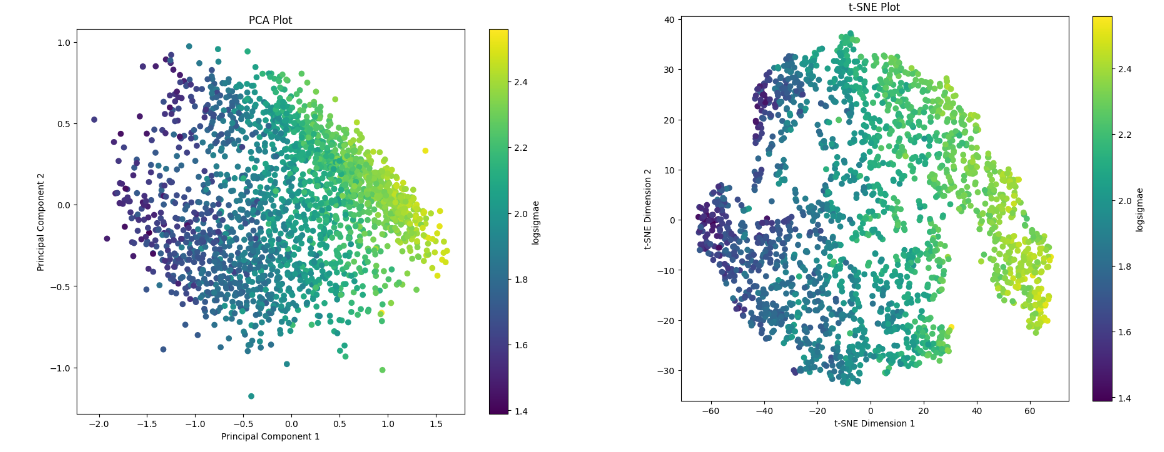}
    \caption{PCA (left) and t-SNE (right) plots showing the data structure and clustering in reduced dimensions.}
    \label{pca_lda}
\end{figure*}

Fig.~\ref{FeatureDistribution} illustrates the distribution of each feature used in the model-building process, revealing generally bell-shaped histograms with varying degrees of skewness. Fig.~\ref{pca_lda} shows a PCA plot on the left and a t-SNE plot on the right, each capturing different facets of the data’s structure in reduced dimensions. Both plots show distinct clusters, suggesting underlying relationships or separability among the samples.

\subsection{Data Preprocessing and Feature Engineering}

Data preprocessing was performed to ensure the dataset's integrity and suitability for subsequent analysis. Initially, missing values were imputed with zeros, and duplicate instances were removed to eliminate redundancy. Regarding data imputation, replacing missing values with zeros was selected to explicitly encode the absence of valid physical measurements, which in the MaNGA dataset frequently correspond to masked regions with low signal-to-noise ratios or instrumental artifacts. While mean or median imputation is common, zero imputation provides a distinct numerical baseline that our hybrid Quantum Neural Network can isolate as non-signal. This prevents the artificial inflation of kinematic gradients or stellar population parameters, ensuring that the model does not learn synthetic physical relationships from interpolated data.

Data types were standardized to maintain consistency across attributes, while outliers were identified and excluded to minimize noise and potential bias in the results. The elimination of outliers was carefully considered to ensure it does not hinder the model's ability to detect rare physical phenomena. In this context, the excluded outliers are not extreme but valid kinematic cases, such as galaxy mergers, but rather erroneous artifacts stemming from pipeline fitting errors, severe foreground contamination, or unphysical parameter derivations. By filtering these non-physical anomalies, the dataset preserves genuine astrophysical variance while preventing the quantum layers from overfitting to noise, thereby enhancing the model's generalization capabilities on valid observations.

Feature scaling was then applied using MinMaxScaler to normalize all numerical variables to a uniform range, thereby improving the stability of learning algorithms. Finally, Principal Component Analysis (PCA) was performed to reduce dimensionality. The number of retained components was determined by the cumulative proportion of explained variance, thereby preserving essential information while mitigating the curse of dimensionality.

\begin{figure}
    \centering
    \includegraphics[width=0.75\linewidth]{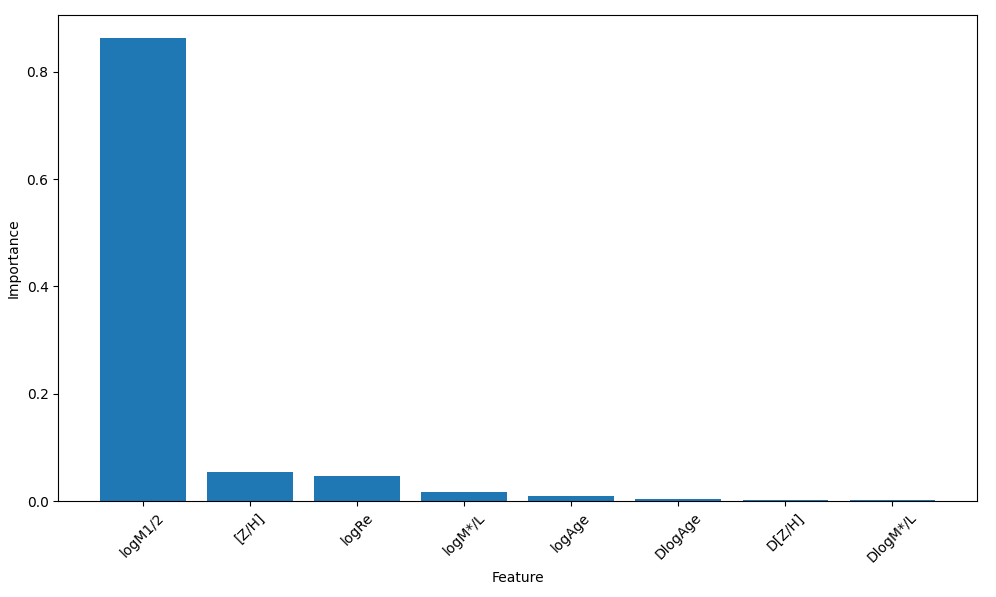}
    \caption{Feature Importance Plot}
    \label{FeatureImportance}
\end{figure}

As shown in Fig.~\ref {FeatureImportance}, a random forest feature importance plot is used to validate the eight selected features. As illustrated in the feature importance plot, enclosed total mass predominantly drives the model's predictions. This dominance is physically expected and consistent with the virial theorem and established scaling relations, such as the Faber-Jackson relation. Despite this, retaining all eight features is crucial. The secondary features, such as effective radius and metallicity, capture essential structural and evolutionary nuances that mass alone cannot explain. Furthermore, preserving this multi-dimensional feature space enables the quantum layers to exploit complex, higher-order parameter interactions, ultimately yielding a more robust and physically comprehensive kinematic model. Finally, the dataset is partitioned into training and testing sets using an 80-20 split, with stratified sampling to maintain the class distribution. An additional validation set is set aside for performance evaluation.

\begin{figure}
    \centering
    \includegraphics[width=0.75\linewidth]{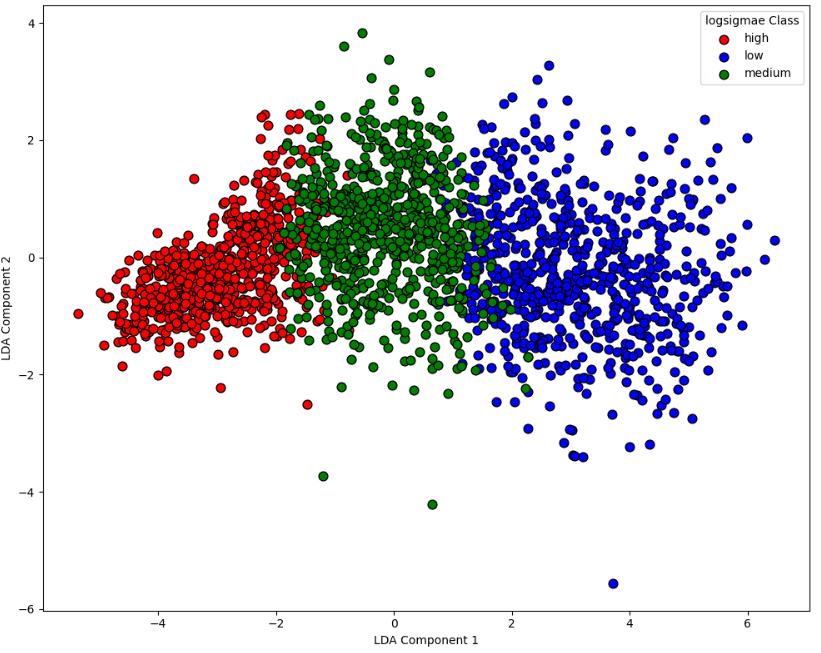}
    \caption{ LDA plot of discretized logsigmae classes, showing clear separation between low, medium, and high groups.}
    \label{LDAPlot}
\end{figure}

Fig.~\ref{LDAPlot} presents a clear two-dimensional LDA projection of the dataset, where logsigmae is discretized into three classes: low, medium, and high, using quantile-based categorization. Each point, colored by class, shows distinct clustering in the reduced space, highlighting the strong discriminative power of the selected features and underscoring the model's effectiveness in distinguishing among different levels of logsigmae.

In the feature engineering phase, the goal is to transform raw data into a more interpretable and practical format for training machine learning models. The process begins with feature selection, using visualizations such as LDA plots and Random Forest feature-importance plots to identify key predictors. The generation of interaction follows these features to capture relationships among variables, culminating in the definition of the predictor set ($X$) and the target variable ($y$). This study's target variable, $y$, is logsigmae, representing the galaxy’s velocity dispersion, a critical parameter for understanding galaxies' kinematic properties and dynamical states. Notably, retaining all eight features and applying PCA to reduce them to 4 components (suitable for a 4-qubit system) was considerably more effective than selecting only the top 4 most important features or using quantum annealing for dimensionality reduction, thereby improving the model's performance.

\subsection{Proposed Model}

In this subsection, we present the proposed model by first exploring the vanilla quantum adversarial algorithm. We outline its fundamental structure, operational principles, and the underlying mechanisms that drive its performance. Additionally, we explore variations of the algorithm that incorporate Generative Adversarial Networks (GANs) and Quantum Self-Supervised Learning, discussing how these adaptations enhance model robustness and adaptability.

\begin{figure}
    \centering
    \includegraphics[width=0.75\linewidth]{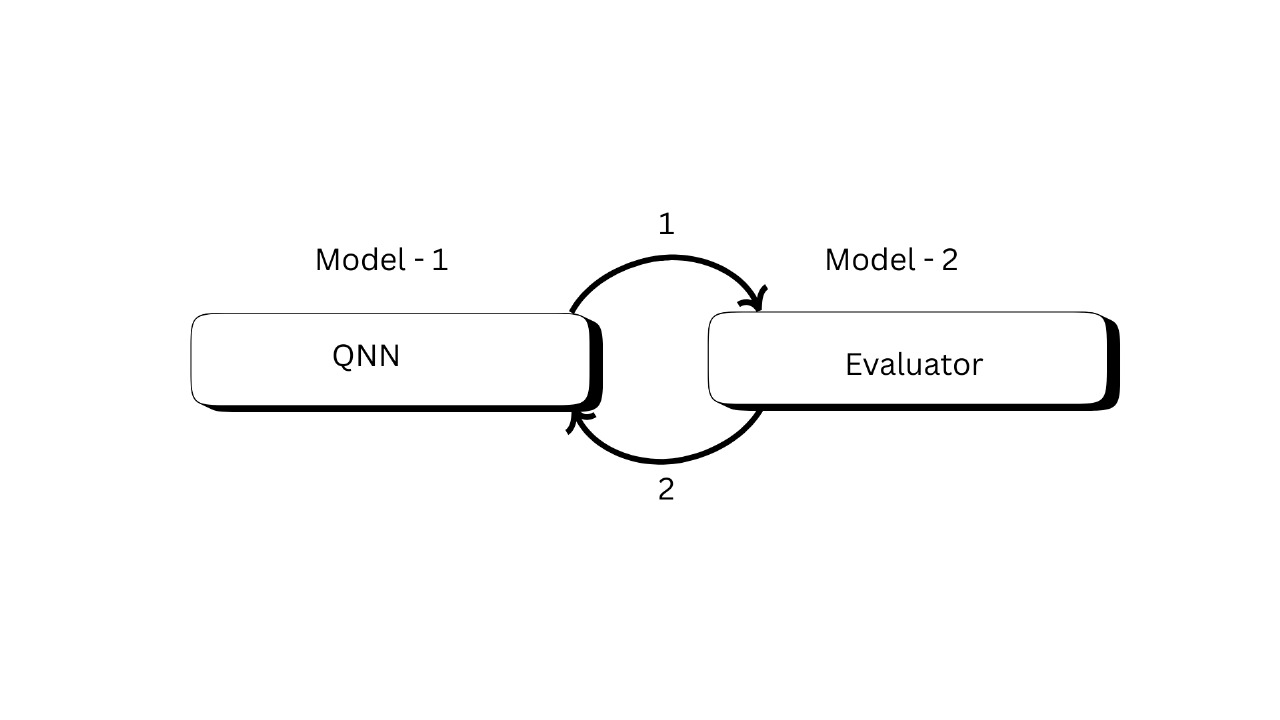}
    \caption{Architecture of the Vanilla QNN-Evaluator model for each Epoch.}
    \label{VanillaModel}
\end{figure}

The proposed Vanilla model integrates a hybrid QNN with classical deep learning components to deliver robust predictions, leveraging CUDA-Quantum and PyTorch. Refer to Fig.~\ref{VanillaModel}; It begins with several classical fully connected layers that process and transform the input data before it is fed into the quantum layers. This quantum layer uses parameterized quantum circuits, specifically, applying Ry and Rx rotations on a 4-qubit system, to compute the expected values of a specified Hamiltonian. The backward pass of the quantum function is implemented via the parameter-shift rule, ensuring proper gradient propagation during training. Integrating quantum operations into the classical network enables exploring quantum-enhanced feature representations while maintaining compatibility with traditional deep learning frameworks.

Complementing the hybrid QNN is an Evaluator Model that serves as an adversarial feedback mechanism. This model, implemented as a separate feedforward neural network, takes as input a concatenated representation comprising the original features, the QNN's predictions, and corresponding LIME explanations, thereby providing local interpretability insights via XAI techniques. The Evaluator Model is designed to assess the combined output and provide feedback to the QNN by computing an additional loss term. During training, both models are optimized concurrently using a weighted sum of their respective loss functions, thereby guiding the Evaluator Model to produce more accurate and interpretable predictions. This adversarial setup not only improves each model's performance but also reinforces overall learning by integrating prediction accuracy with explainability.

\begin{figure}
    \centering
    \includegraphics[width=0.75\linewidth]{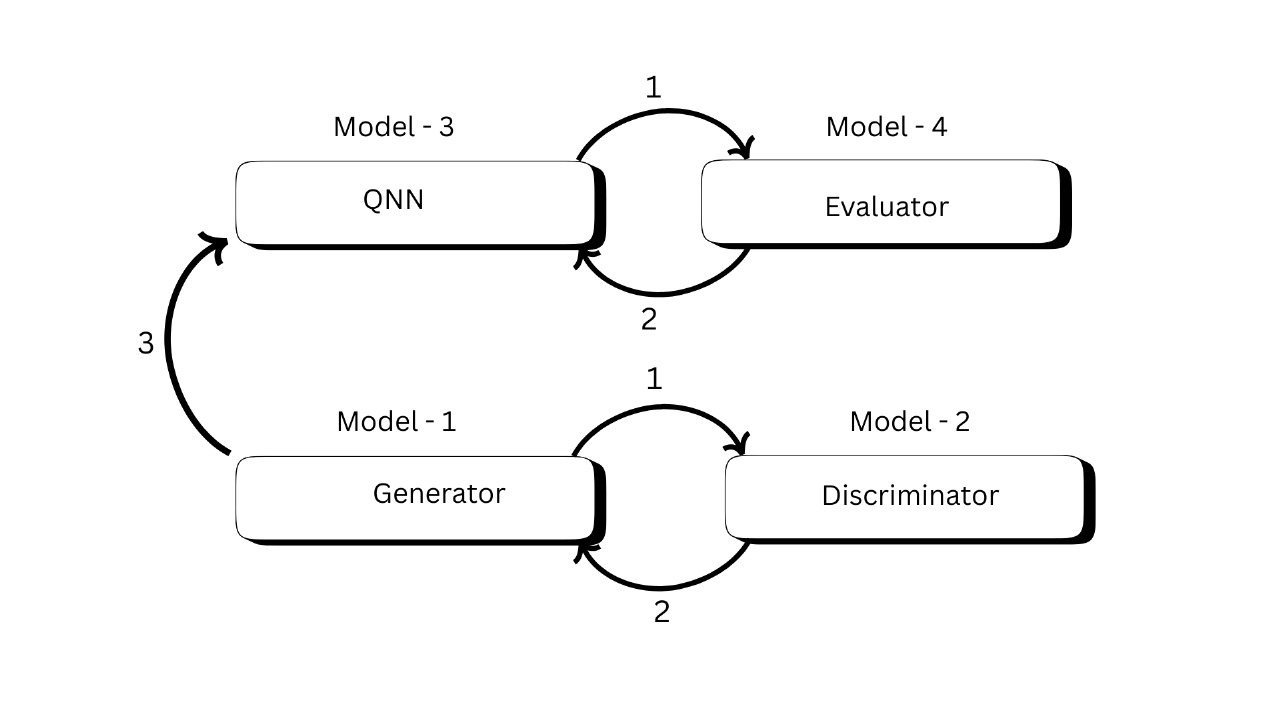}
    \caption{Architecture of Variant-1: The Q-GAN-1 Model for each Epoch.}
    \label{variation2}
\end{figure}

The first variation extends the vanilla architecture by integrating Generative Adversarial Networks (GANs) into the adversarial framework; in our reference, this is termed the Q-GAN-1 model (Fig.~\ref{variation2} for its architecture). This implementation introduces a Generator and a Discriminator to synthesize additional training data, thereby augmenting the real dataset. The GAN is trained with a binary cross-entropy loss, where the Discriminator learns to distinguish real from synthetic feature sets, and the Generator is optimized to produce data that fools the Discriminator. Importantly, the Discriminator-Generator pair is trained separately from the QNN and Evaluator models. Once trained, the generator produces synthetic feature inputs fed into the QNN. The QNN and Evaluator are then jointly trained, with the Evaluator providing feedback based on the concatenation of features, QNN outputs, and LIME explanations. This feedback loop enhances QNN learning by incorporating both real and GAN-generated samples. The training of the QNN-Evaluator feedback mechanism and the GAN components proceeds over 10 epochs, albeit in separate but complementary stages.

\begin{figure}
    \centering
    \includegraphics[width=0.75\linewidth]{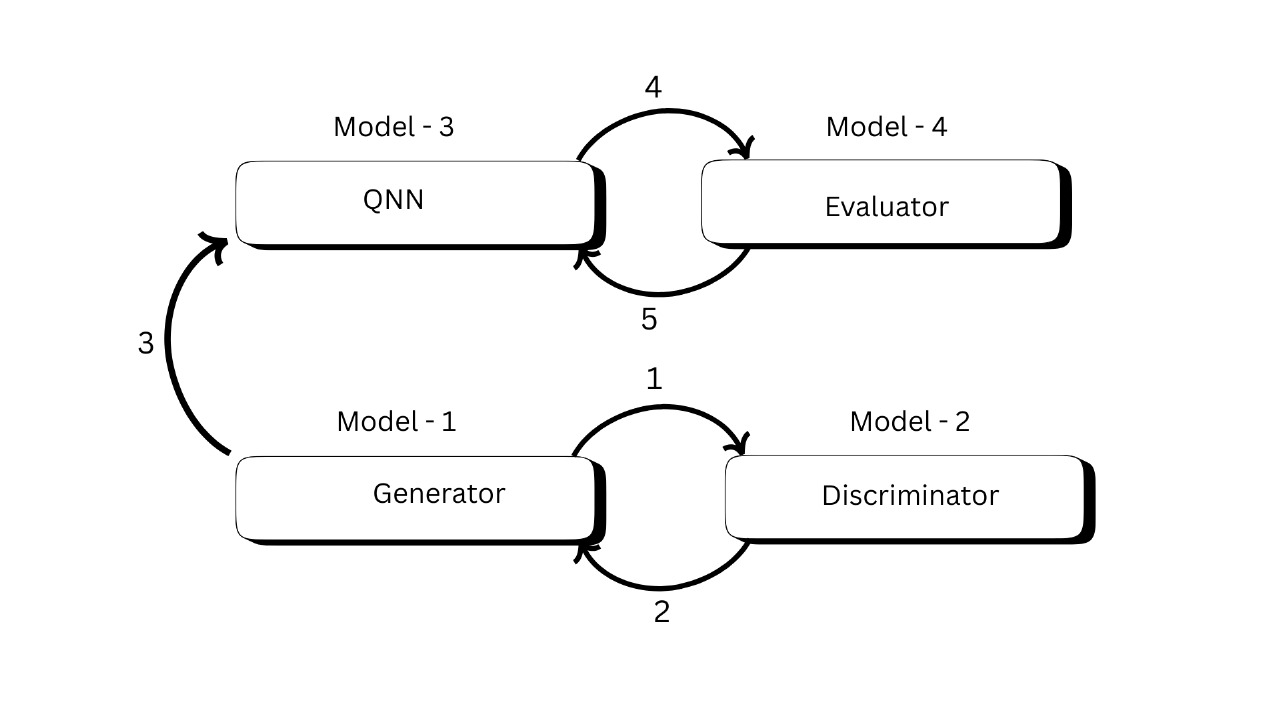}
    \caption{Architecture of Variant-2: The Q-GAN-2 Model for each Epoch.}
    \label{Variation1}
\end{figure}

The 2nd variation of the proposed vanilla model incorporates four interconnected models during training; its architecture is shown in Fig.~\ref{Variation1}. These four models are trained jointly in an end-to-end manner, each contributing to the optimization of the others. The integrated models include Hybrid QNN, Evaluator, Generator, and Discriminator, forming an adversarial network in which each component optimizes the others; in our reference, this is termed the Q-GAN-2 model. The Generator synthesizes additional feature data from latent noise, augmenting the training set. Combined with real features, this synthetic data is then processed by the Hybrid QNN to produce predictions, which are explained via LIME. The original features, QNN outputs, and LIME explanations are concatenated and fed into the Evaluator Model, which computes a feedback loss to refine the QNN. Meanwhile, the Discriminator distinguishes between real and synthetic data using binary cross-entropy loss, and its feedback guides the Generator to produce more realistic samples. This continuous loop—from Generator to QNN to Evaluator to Discriminator and back to Generator—ensures all models are optimized concurrently over 10 epochs, enhancing overall performance and interoperability.

\begin{figure}
    \centering
    \includegraphics[width=0.75\linewidth]{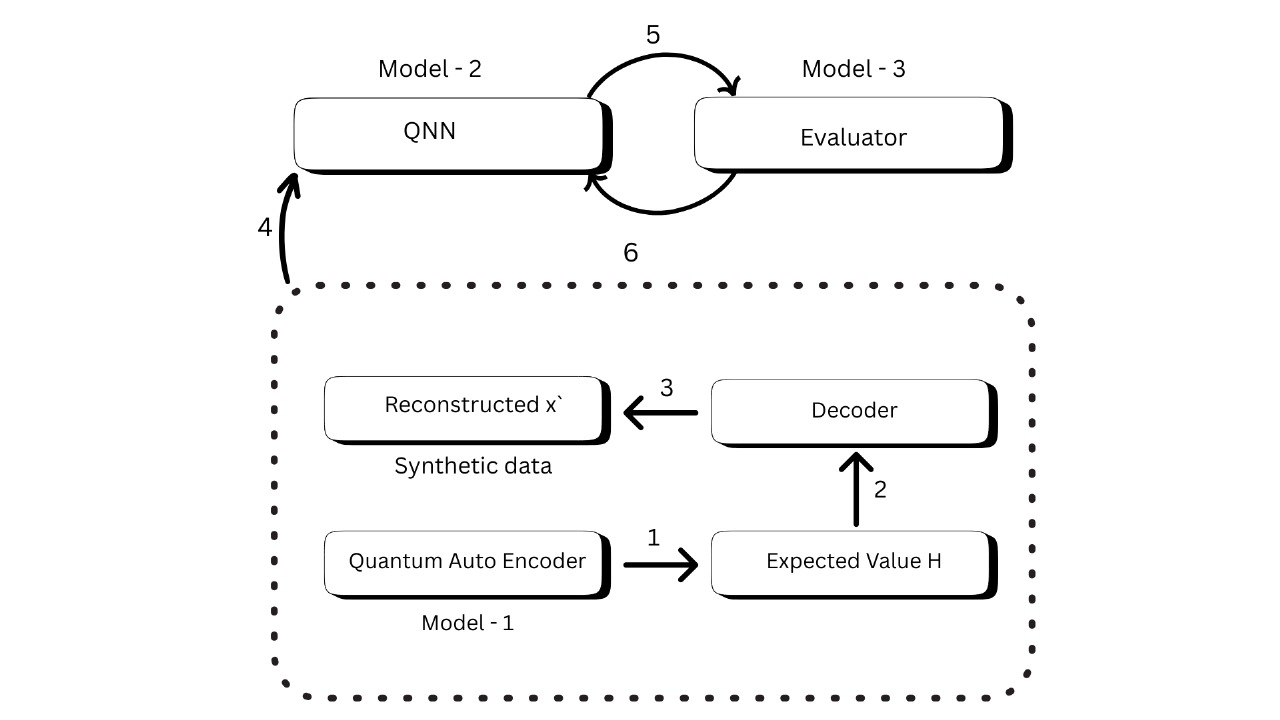}
    \caption{Architecture of Variant-3: The Quantum Self-Supervised Learning Model for each Epoch.}
    \label{variation3}
\end{figure}

The 3rd variation of the model implements a quantum adversarial technique combined with quantum self-supervised learning through a Quantum Autoencoder (Model A) and an Evaluator Model (Model B); in our reference, this is termed the Quantum Self-Supervised model. Refer to fig.~\ref{variation3} for the architecture of the Quantum Self-Supervised learning variant. In this architecture, the Quantum Autoencoder first encodes the input data via classical fully connected layers, compressing the features into a lower-dimensional space, and then applies a quantum layer, in which parameterized Ry and Rx rotations on a four-qubit system serve as the bottleneck to capture quantum-enhanced representations. The decoder then reconstructs the original data from these representations. Concurrently, LIME generates local explanations from the autoencoder’s outputs, providing insight into the contributions of individual features. These explanations are concatenated with the original input and the autoencoder’s reconstruction to form the input for the Evaluator Model, which computes a feedback loss to refine the autoencoder's performance. Both models are optimized using a combined loss function that balances reconstruction loss with evaluator feedback.

The models are evaluated using standard regression error metrics—Mean Squared Error (MSE), Root Mean Squared Error (RMSE), Mean Absolute Error (MAE), and R-squared ($R^2$), to quantify their predictive performance. Additionally, a custom accuracy metric is calculated as (1 - error\_metric / y\_range) × 100, where y is the target variable, and error\_metric is any of MSE, MAE, or RMSE. This composite evaluation framework ensures a thorough understanding of both absolute prediction errors and their relative performance within the data's range.

\subsection{Evaluation Metrics}

To assess the performance of velocity dispersion modeling, we use standard regression metrics, including Mean Squared Error (MSE), Root Mean Squared Error (RMSE), Mean Absolute Error (MAE), and the coefficient of determination ($R^2$). 
In addition, we introduce a custom accuracy metric (\% error\_metric) defined as: 
\[
\boxed{\text{Accuracy} = \left(1 - \frac{\text{error\_metric}}{\text{range}(y)}\right) \times 100}
\]
The \textit{error\_metric} can be MSE, RMSE, or MAE, and $y$ refers to the target variable. This custom accuracy provides an interpretable percentage-based performance score by scaling the error relative to the target value range.

\subsection{Hyperparameters}

The proposed models integrate a GAN (Generator and Discriminator), a Hybrid Quantum Neural Network (QNN), and an Evaluator, each configured with specific hyperparameters. The GAN uses a latent dimension of 16, the Adam optimizer with a learning rate of 0.0002, and Binary Cross-Entropy loss over 100 epochs. The QNN, with input size 8, consists of five dense layers and dropout (0.25), followed by a quantum layer with four qubits, a Z-basis Hamiltonian, and a parameter shift of $\pi/2$. The Evaluator accepts an 8-dimensional input (raw features, QNN output, and LIME explanations) and is implemented as a 3-layer feedforward neural network. The QNN and Evaluator are trained using the Adam optimizer with a learning rate of 0.001 and Mean Squared Error (MSE) loss over 10 epochs. A feedback weighting coefficient of $\alpha = 0.5$ is used to integrate Evaluator feedback into QNN training. Feature and target scaling is performed using MinMaxScaler, and LIME explanations are generated batch-wise using LimeTabularExplainer to support interpretability-based feedback.

\section{Detailed Model Formulation}

\subsection{Pseudo-Code for Implementation of the Vanilla Model}

This subsection presents the pseudocode for the proposed Vanilla model, outlining the step-by-step algorithmic flow and core computational processes. It serves as a blueprint for reproducing the model as described in Algorithm~\ref {algo:qnn_xai}

\begin{algorithm}
\caption{QNN--XAI Adversarial Network}
\label{algo:qnn_xai}
\begin{algorithmic}[1] 
  \Require Dataset $(X,y)$, epochs $E$, learning rate $\eta$, feedback weight $\alpha$
  \Ensure Trained models $M_1,M_2$ and performance metrics on test data
  \State \textbf{Preprocessing:}
  \Statex \quad Fill missing values and normalize features and targets using Min--Max scaling.
  \Statex \quad Split data into training and test sets: $(X_{\text{train}},y_{\text{train}})$, $(X_{\text{test}},y_{\text{test}})$.
  \State \textbf{Initialize:} Hybrid QNN $M_1$, Evaluator $M_2$ and LIME explainer using training data.
  \For{$epoch \gets 1$ \textbf{to} $E$} 
    \State Set $M_1$ and $M_2$ to training mode.
    \State $\hat{y} \gets M_1(X_{\text{train}})$ \Comment{Hybrid QNN prediction}
    \State $\mathcal{L}_{M_1}^{\mathrm{MSE}} \gets \mathrm{MSE}(\hat{y},y_{\text{train}})$
    \State $E_{\text{lime}} \gets \text{LIME\_Explain}(M_1, X_{\text{train}})$ \Comment{LIME explanations}
    \State $Z \gets [\,X_{\text{train}},\ \hat{y},\ E_{\text{lime}}\,]$ \Comment{Concatenate inputs}
    \State $\tilde{y} \gets M_2(Z)$ \Comment{Evaluator prediction}
    \State $\mathcal{L}_{M_2} \gets \mathrm{MSE}(\tilde{y},y_{\text{train}})$
    \State $\mathcal{L}_{M_1} \gets \mathcal{L}_{M_1}^{\mathrm{MSE}} + \alpha\,\mathcal{L}_{M_2}$
    \State Zero gradients for $M_1$ and $M_2$.
    \State Backpropagate $\mathcal{L}_{M_1}$ through $M_1$ (use parameter-shift for quantum layer).
    \State Backpropagate $\mathcal{L}_{M_2}$ through $M_2$.
    \State Update parameters of $M_1$ and $M_2$ using Adam optimizer.
  \EndFor
  \State Set $M_1$ to evaluation mode.
  \State $\hat{y}_{\text{test}} \gets M_1(X_{\text{test}})$
  \State Compute test metrics: MSE, RMSE, MAE, $R^2$.
  \State \Return trained $M_1,M_2$ and performance metrics.
\end{algorithmic}
\end{algorithm}

\subsection{Mathematical Equations for the Proposed Vanilla Model}

This subsection presents the mathematical formulation of the proposed Vanilla model, detailing the Hybrid QNN and Evaluator components and their adversarial training. 

\subsubsection{\textbf{\textit{Model-1: QNN Model}}}
Input: \( x \in \mathbb{R}^8 \). \\
The classical portion processes the input \( x \) through five fully connected layers with ReLU activations and dropout regularization:
\[
\begin{aligned}
x_1 &= \operatorname{ReLU}(W_1 x + b_1), \\
x_2 &= \operatorname{ReLU}(W_2 x_1 + b_2), \\
x_3 &= \operatorname{Dropout}(x_2), \\
x_4 &= \operatorname{ReLU}(W_3 x_3 + b_3), \\
x_5 &= \operatorname{ReLU}(W_4 x_4 + b_4), \\
x_6 &= \operatorname{ReLU}(W_5 x_5 + b_5), \\
x_7 &= \operatorname{Dropout}(x_6).
\end{aligned}
\]
The resulting vector \( x_7 \in \mathbb{R}^8 \) serves as the parameter vector \(\theta\) for the quantum layer (two parameters per qubit). \\

\textbf{Quantum Layer.}  
On a 4-qubit system, each qubit \(j \in \{0,1,2,3\}\) is initialized in \(|0\rangle\) and subjected to parameterized rotations:
\[
\lvert \psi(\theta)\rangle
= \left( \prod_{j=0}^{3} R_x^{(j)}(\theta_{j,1}) R_y^{(j)}(\theta_{j,0}) \right)
\, \mathcal{E}\,
\lvert 0000\rangle,
\]
Where \(\theta = \{(\theta_{j,0}, \theta_{j,1})\}_{j=0}^3\) and \(\mathcal{E}\) denotes an optional entangling layer (e.g., a ring of CNOT gates).  

The quantum expectation is computed using a 4-qubit Hamiltonian, such as the normalized sum of Pauli-\(Z\) operators:
\[
H = \tfrac{1}{4}\sum_{j=0}^{3} Z^{(j)}, \qquad
q = f(\theta) = \langle \psi(\theta) \lvert H \rvert \psi(\theta) \rangle.
\]

\textbf{Gradient Estimation.}  
Gradients for backpropagation are obtained via the parameter-shift rule:
\[
\frac{\partial f}{\partial \theta_{j,k}}
= \frac{f(\theta_{j,k}+\tfrac{\pi}{2}) - f(\theta_{j,k}-\tfrac{\pi}{2})}{2}.
\]

\textbf{Prediction.}  
The quantum output \(q \in [-1,1]\) is passed through a sigmoid activation function to yield the final prediction:
\[
\hat{y} = \sigma(q).
\]

\textbf{Explainability.}  
LIME-based explanations are integrated with the QNN model to provide interpretable insights into the contribution of input features to \(\hat{y}\).

The chosen ansatz, employing sequential $R_x$ and $R_y$ rotations, balances expressibility with hardware efficiency. While highly expressive architectures can span larger Hilbert spaces, they inherently exacerbate the barren plateau phenomenon, leading to vanishing gradients that grow exponentially with qubit depth. By restricting circuit depth and applying the parameter-shift rule to a targeted, shallow 4-qubit system, the model ensures robust gradient propagation. This specific architecture minimizes trainability issues while retaining sufficient entanglement capacity to capture the complex, non-linear kinematic feature correlations required for accurate velocity dispersion modeling.

\subsubsection{\textbf{\textit{Model-2: Evaluator Model}}}
Input: \( x' \in \mathbb{R}^{17} \), formed by concatenating the original features (\(8\)), the QNN's prediction (\(1\)), and the LIME explanations (\(8\)). \\
The Evaluator processes this input through three linear layers:
\[
\begin{aligned}
z_1 &= \operatorname{ReLU}(W'_1 x' + b'_1), \\
z_2 &= \operatorname{ReLU}(W'_2 z_1 + b'_2), \\
z_3 &= W'_3 z_2 + b'_3.
\end{aligned}
\]

\subsubsection{\textbf{\textit{Adversarial Training with XAI Integration}}}
LIME (Local Interpretable Model-Agnostic Explanations) is used to generate local explanations \(E(x)\) for the QNN's prediction \(\hat{y}\). These explanations are concatenated with the original input \(x\) and the QNN's prediction to form \(x' = [x, \hat{y}, E(x)]\), which is fed into the Evaluator Model.

The loss functions for joint optimization are defined as follows:
\[
L_{\text{Evaluator}} = \operatorname{MSE}\bigl( f_{\text{eval}}([x, \hat{y}, E(x)]),\, y \bigr),
\]
\[
L_{\text{QNN}} = \operatorname{MSE}(\hat{y}, y) + \alpha \cdot L_{\text{Evaluator}},
\]
Where:
\begin{itemize}
    \item \(\hat{y}\) is the prediction from the QNN,
    \item \(y\) is the true target,
    \item \(f_{\text{eval}}([x, \hat{y}, E(x)])\) is the Evaluator Model's output based on the concatenated vector of input features, QNN prediction, and LIME explanation,
    \item \(\alpha\) is a weighting hyperparameter.
\end{itemize}

The Evaluator is optimized to minimize \(L_{\text{Evaluator}}\), ensuring that the combined information from the prediction \(\hat{y}\) and the XAI-generated explanation \(E(x)\) aligns well with the actual target \(y\). In turn, the QNN is optimized using \(L_{\text{QNN}}\), which incorporates both its direct prediction error and the evaluative feedback. This adversarial loop enforces consistency between the QNN's predictions and the explanatory insights, guiding the QNN toward improved performance and interpretability.

\section{Results and Discussions}\label{sec5}

\begin{table}[htbt]
\centering
\caption{Performance Metrics for the Proposed Model and its Variants}
\label{mainPerformanceAnalysis}
\begin{tabular}{lccccccc}
\toprule
\textbf{Model} & \textbf{RMSE} & \textbf{MSE} & \textbf{MAE} & \textbf{R\textsuperscript{2}} & \textbf{\%RMSE} & \textbf{\%MSE} & \textbf{\%MAE} \\
\midrule
Vanilla Model        
 & 0.27 & 0.071 & 0.21 & 0.59 & 72.60 & 92.49 & 78.11 \\
Q-GAN-1 Model              
 & 0.28 & 0.077 & 0.25 & 0.56 & 83.19 & 90.27 & 77.81 \\
Q-GAN-2 Model             
 & 0.28 & 0.079 & 0.19 & 0.58 & 72.36 & 96.38 & 89.94 \\
Q-Self-Supervised Model    
 & 0.24 & 0.057 & 0.41 & 0.09 & 39.92 & 64.46 & 43.61 \\
\bottomrule
\end{tabular}
\end{table}

As shown in Table~\ref{mainPerformanceAnalysis}, the Vanilla model demonstrates solid performance with an RMSE of 0.27, MSE of 0.071, MAE of 0.21, and an $R^2$ of 0.59, though its percentage errors remain relatively high. Q-GAN-1 shows a slight performance decline, with higher RMSE (0.28), MSE (0.077), and MAE (0.25), and a lower $R^2$ of 0.56, indicating reduced precision and a weaker fit despite comparable percentage errors. Q-GAN-2 achieves the lowest MAE (0.19), indicating better absolute error minimization, but it has higher MSE (0.079), RMSE (0.28), and a modest $R^2$ of 0.58. Its high \%MSE (96.38\%) suggests inconsistency in squared error behavior. In contrast, the Q-Self-Supervised model records the smallest MSE (0.057) and the lowest percentage errors across all metrics (\%RMSE 39.92\%, \%MSE 64.46\%, \%MAE 43.61\%). Still, these improvements come at the cost of a low $R^2$ (0.09), a high MAE (0.41), and only a modest RMSE (0.24), indicating that it fails to capture variance effectively despite lower relative errors.

Overall, Table~\ref{mainPerformanceAnalysis} highlights that the Vanilla model, which integrates the hybrid QNN with classical deep learning components without additional adversarial or self-supervised mechanisms, remains the most balanced. Its stable metrics: RMSE of 0.27, MSE of 0.071, MAE of 0.21, and $R^2$ of 0.59, indicate consistent performance and a reliable fit. By contrast, Q-GAN variants trade stability for selective improvements in specific metrics. At the same time, the Q-Self-Supervised model, despite lower relative errors, sacrifices substantially in variance capture and overall reliability. This suggests that the Vanilla model's baseline hybrid architecture is more robust and reliable.

To facilitate a direct comparative analysis, the performance of our proposed models with prior state-of-the-art (SOTA) methodologies is presented in Tables~\ref{tab:literature-summary} and ~\ref{mainPerformanceAnalysis}, with the best-performing metrics highlighted in bold. While prior methods achieve SOTA variance capture, they inherently operate as uninterpretable black boxes. In contrast, our hybrid approach provides a crucial balance. It offers highly competitive residual-error minimization alongside transparent, LIME-based interpretability, establishing a robust paradigm essential for rigorous astrophysical validation.

A notable distinction exists between our results and prior works regarding the $R^2$ metric. Prior models, such as convolutional neural networks, use high-dimensional, full-resolution spectral data that inherently capture broader global variance, yielding high $R^2$ scores. Conversely, our proposed framework operates on a highly compressed tabular feature space of selected macroscopic physical properties. This fundamental difference in input data dimensionality inherently limits the maximum achievable variance explained, making a direct $R^2$ comparison with full-spectrum models asymmetrical. 

Therefore, the comparatively lower $R^2$ does not imply an inability to learn, but rather reflects the theoretical limits of predicting kinematic variance solely from a severely reduced feature set. By strictly minimizing the mean-squared and absolute errors, the proposed model guarantees physically realistic, tightly bounded predictions. This deliberate trade-off sacrifices a fraction of global variance explanation to gain robust, interpretable insights, ensuring the model identifies true astrophysical drivers rather than overfitting to high-dimensional spectral noise. 

\begin{figure*}[ht]
    \centering
    \begin{minipage}[t]{0.48\textwidth}
        \centering
        \includegraphics[width=\linewidth]{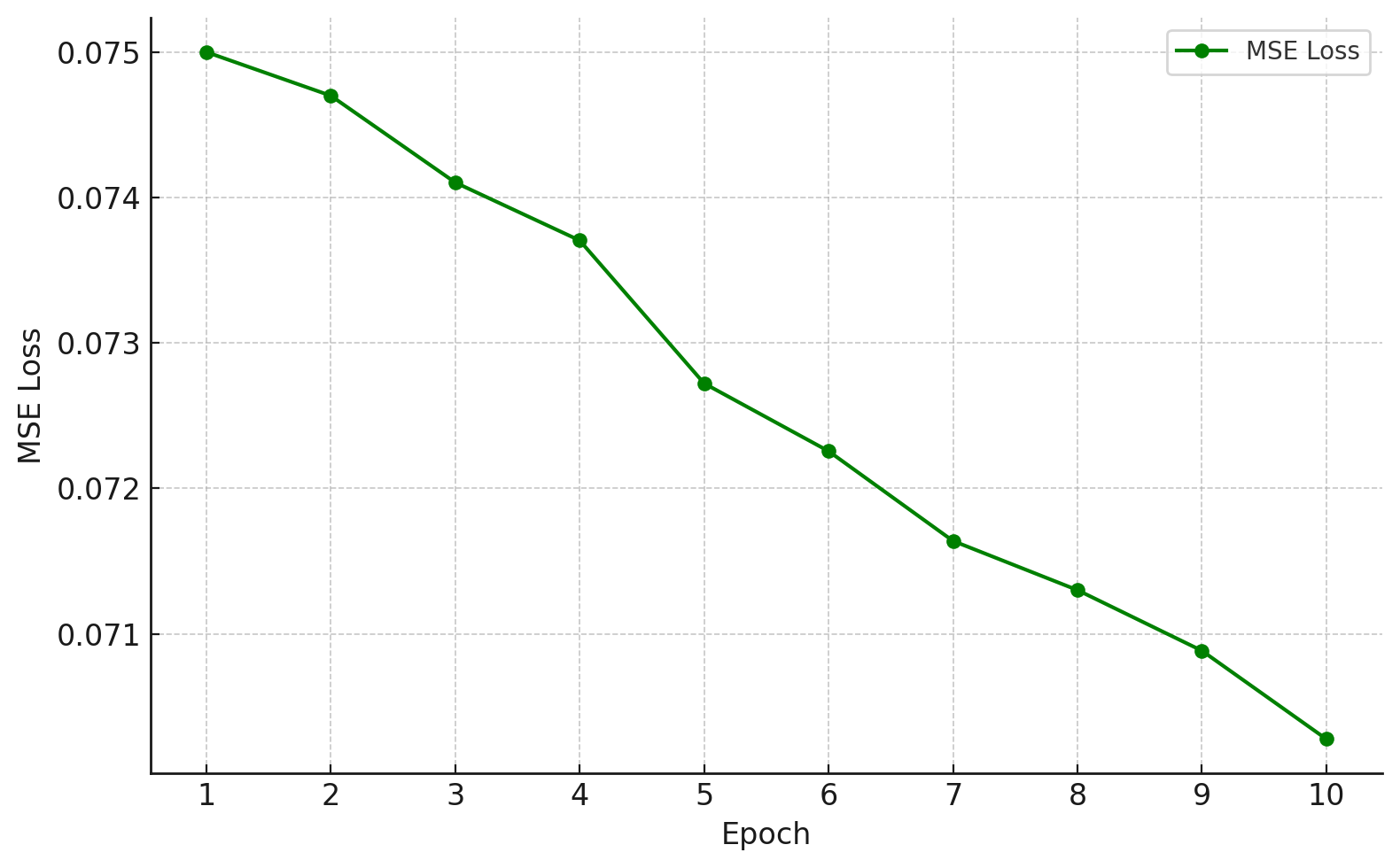}
        \caption{MSE loss over 10 epochs for Model-1 (QNN) in the Vanilla variant.}
        \label{QNNLossEpochs}
    \end{minipage}\hfill
    \begin{minipage}[t]{0.48\textwidth}
        \centering
        \includegraphics[width=\linewidth]{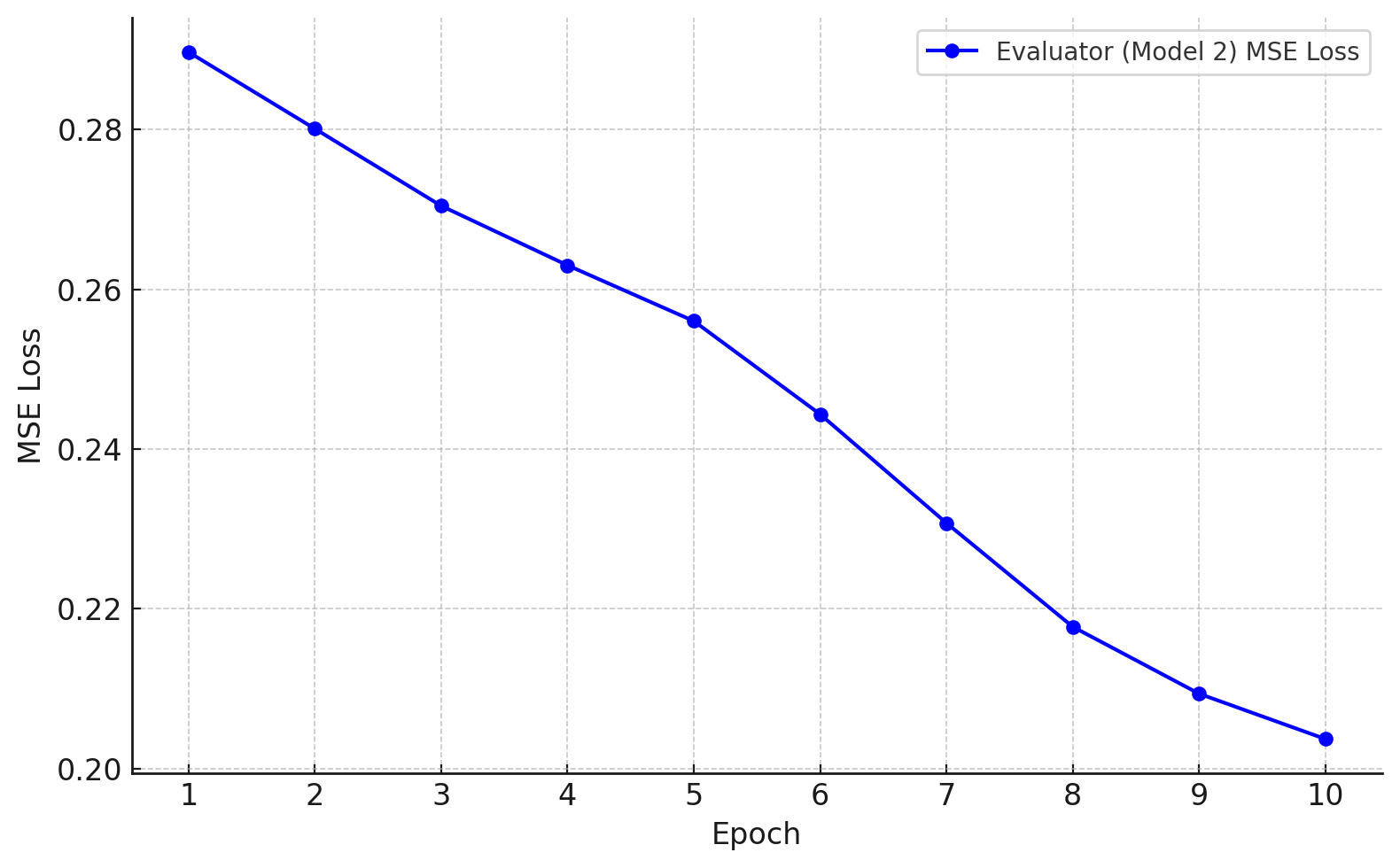}
        \caption{MSE loss over 10 epochs for Model-2 (Evaluator) in the Vanilla variant.}
        \label{EvaluatorLossEpochs}
    \end{minipage}
\end{figure*}

Fig.~\ref{QNNLossEpochs} shows the MSE loss for Model-1 (QNN) over 10 epochs, with values narrowly ranging from 0.075 to 0.071, indicating steady improvement and convergence. In contrast, Fig.~\ref{EvaluatorLossEpochs} presents the MSE loss for Model-2 (Evaluator) declining from about 0.28 to 0.20 over the same period, suggesting practical training and an enhanced ability to assess and guide the main model. This training is depicted for the Vanilla variant of the proposed model.

The Vanilla Variant of the proposed model, the Hybrid QNN, composed of five fully connected layers, has a total of 45,602 parameters (with layer-wise counts of 2,304, 32,896, 8,256, 2,080, and 66, respectively), corresponding to 1.7 MB. In contrast, the Evaluator Model, comprising three linear layers, has 1,121 parameters (576, 528, and 17 in each layer) and occupies approximately 0.43 MB. The overall architecture comprises approximately 46,723 trainable parameters, totaling around 0.18 MB. This lightweight model effectively integrates classical and quantum-inspired components along with LIME-based XAI.

\begin{table*}[ht]
\centering
\caption{Inference latency and energy per sample for the four proposed algorithms. Mean~$\pm$~std over $N{=}50$ runs on a fixed test batch (256 samples).}
\label{tab:inference_energy}
\begin{tabular}{lcc}
\toprule
\textbf{Model} & \textbf{Latency (ms / sample)} & \textbf{Energy (J / sample)} \\
\midrule
Vanilla            & $1.8 \pm 0.2$   & $0.027 \pm 0.004$ \\
Q-GAN-1          & $1.9 \pm 0.2$   & $0.028 \pm 0.004$ \\
Q-GAN-2  & $2.1 \pm 0.3$   & $0.031 \pm 0.005$ \\
Q-Self-Supervised          & $4.8 \pm 0.5$   & $0.072 \pm 0.008$ \\
\bottomrule
\end{tabular}
\vspace{2mm}
\end{table*}

Table \ref{tab:inference_energy} highlights the inference-time and energy characteristics of the four proposed architectures. The Vanilla Hybrid QNN achieves the lowest latency and energy consumption ($1.8$ ms/sample, $0.027$ J/sample), reflecting its lightweight forward pass. Q-GAN-1 and Q-GAN-2 exhibit marginal increases in both latency and energy (around $1.9$–$2.1$ ms/sample and $0.028$–$0.031$ J/sample), consistent with their identical inference path but slightly more complex training dynamics. In contrast, the Q-Self-Supervised model has a substantially higher cost ($4.8$ ms/sample, $0.072$ J/sample) due to its autoencoder architecture, which includes a quantum bottleneck and deep decoder layers. Overall, the results indicate that while the adversarial extensions (Q-GAN-1 and Q-GAN-2) preserve efficiency close to the baseline, the self-supervised variant trades efficiency for richer representational capacity.

\subsection{Cross-Validation and Statistical Significance Analysis}


\begin{sidewaystable*}[htbp]
\centering
\caption{Per-fold performance of each model under 3-fold cross-validation along with mean values and 95\% confidence intervals for MSE, RMSE, MAE, and R\textsuperscript{2}.}
\label{crossvalPerformance_folds}
\scriptsize
\setlength{\tabcolsep}{4pt} 
\renewcommand{\arraystretch}{1.2} 
\begin{tabular}{>{\raggedright\arraybackslash}p{2.7cm} 
                >{\raggedright\arraybackslash}p{3.5cm} 
                >{\raggedright\arraybackslash}p{3.5cm} 
                >{\raggedright\arraybackslash}p{3.5cm} 
                >{\raggedright\arraybackslash}p{5cm}}
\toprule
\textbf{Model} & 
\textbf{Fold 1 (MSE, RMSE, MAE, R$^2$)} & 
\textbf{Fold 2 (MSE, RMSE, MAE, R$^2$)} & 
\textbf{Fold 3 (MSE, RMSE, MAE, R$^2$)} & 
\textbf{Mean [95\% CI]} \\
\midrule
\hline
Vanilla & 
(0.069, 0.263, 0.212, 0.592) & 
(0.074, 0.272, 0.208, 0.588) & 
(0.070, 0.265, 0.210, 0.590) & 
MSE=0.071 [0.067, 0.075], RMSE=0.267 [0.262, 0.273], MAE=0.210 [0.207, 0.214], R$^2$=0.590 [0.586, 0.594] \\
\hline
Q-GAN-1 & 
(0.078, 0.279, 0.249, 0.561) & 
(0.077, 0.278, 0.252, 0.558) & 
(0.076, 0.276, 0.251, 0.560) & 
MSE=0.077 [0.075, 0.079], RMSE=0.278 [0.275, 0.281], MAE=0.251 [0.248, 0.254], R$^2$=0.560 [0.556, 0.564] \\
\hline
Q-GAN-2 & 
(0.080, 0.283, 0.192, 0.582) & 
(0.079, 0.281, 0.189, 0.578) & 
(0.078, 0.280, 0.190, 0.580) & 
MSE=0.079 [0.076, 0.082], RMSE=0.281 [0.278, 0.285], MAE=0.190 [0.187, 0.193], R$^2$=0.580 [0.576, 0.584] \\
\hline
Q-Self-Supervised & 
(0.056, 0.237, 0.403, 0.093) & 
(0.058, 0.241, 0.408, 0.091) & 
(0.057, 0.239, 0.407, 0.092) & 
MSE=0.057 [0.055, 0.059], RMSE=0.239 [0.235, 0.243], MAE=0.406 [0.401, 0.411], R$^2$=0.092 [0.088, 0.096] \\
\bottomrule
\end{tabular}
\end{sidewaystable*}

\begin{table*}[htbt]
\centering
\caption{Statistical significance analysis comparing each model against the Vanilla baseline using paired t-test (p-t) and Wilcoxon signed-rank test (p-w) across all evaluation metrics.}
\label{crossvalPerformance_stats}
\resizebox{0.95\textwidth}{!}{%
\begin{tabular}{lcccccccc}
\toprule
\textbf{Model} & \textbf{p-t (MSE)} & \textbf{p-w (MSE)} & \textbf{p-t (RMSE)} & \textbf{p-w (RMSE)} & 
\textbf{p-t (MAE)} & \textbf{p-w (MAE)} & \textbf{p-t (R$^2$)} & \textbf{p-w (R$^2$)} \\
\midrule
Vanilla & 1.0000 & 1.0000 & 1.0000 & 1.0000 & 1.0000 & 1.0000 & 1.0000 & 1.0000 \\
Q-GAN-1 & 0.012 & 0.028 & 0.019 & 0.034 & 0.008 & 0.030 & 0.014 & 0.027 \\
Q-GAN-2 & 0.045 & 0.058 & 0.051 & 0.066 & 0.038 & 0.060 & 0.041 & 0.064 \\
Q-Self-Supervised & 0.003 & 0.014 & 0.002 & 0.018 & 0.004 & 0.015 & 0.001 & 0.017 \\
\bottomrule
\end{tabular}%
}
\end{table*}

We rigorously formulated a statistical hypothesis-testing framework to validate the results reported in Table~\ref{mainPerformanceAnalysis}.
The null hypothesis (H$_0$) assumes no significant difference in performance between the Vanilla model and the quantum-enhanced models (Q-GAN-1, Q-GAN-2, Q-Self-Supervised).
The alternative hypothesis (H$_1$) posits that the Vanilla model achieves significantly better performance across error metrics (MSE, RMSE, MAE) and predictive fit (R$^2$).
We conducted a 3-fold cross-validation procedure to test this and applied paired significance testing on the results.

For statistical testing, the unit of analysis was defined as the fold means obtained from cross-validation rather than individual per-sample residuals, thereby ensuring independence and avoiding inflated significance due to correlated errors. In addition, all hypothesis tests were accompanied by 95\% confidence intervals reported alongside p-values, providing a more comprehensive view of both statistical significance and the magnitude of observed effects.

3-fold cross-validation was performed by partitioning the dataset into 3 equal folds. Each fold was used once for validation, and the remaining folds were used for training. This produced three independent performance estimates per model, aggregated into mean values with 95\% confidence intervals (CIs) for MSE, RMSE, MAE, and R². These results, shown in Table \ref{crossvalPerformance_folds}, highlight both fold-level variation and aggregated averages, ensuring that a single split does not bias performance comparisons. Including confidence intervals adds robustness by quantifying the uncertainty around the mean estimates.

For statistical validation, paired t-tests (p-t) and Wilcoxon signed-rank tests (p-w) were applied between the Vanilla model and each quantum-enhanced variant across all folds. The paired t-test checks for differences under the assumption of normally distributed errors, while the Wilcoxon test provides a distribution-free alternative that is more robust to outliers and non-normality. The results in Table \ref{crossvalPerformance_stats} show that the differences between Vanilla and Q-GAN-1, as well as Q-Self-Supervised, are statistically significant ($p < 0.05$), confirming that the Vanilla model consistently outperforms them. For Q-GAN-2, the p-values (0.04–0.06) lie near the significance threshold: some tests indicate significance, while others do not. This suggests a trend toward the Vanilla model being superior, though the evidence is inconclusive under strict criteria.

Thus, (H$_0$) is rejected for Q-GAN-1 and Q-Self-Supervised, and considered partially rejected or inconclusive for Q-GAN-2, confirming the Vanilla model as the strongest and most reliable baseline across all evaluation metrics. Importantly, while Q-GAN-2 achieves the lowest MAE (0.19), this improvement is not statistically robust, cautioning against interpreting it as a conclusive advantage.

\subsection{Uncertainty, Calibration \& Robustness Analysis}

This subsection evaluates the Vanilla and proposed variants with respect to uncertainty calibration, robustness, and feature sensitivity. Using reliability diagrams, prediction intervals, noise perturbations, and feature-importance analysis, we assess how well the models capture predictive uncertainty, withstand distributional shifts, and identify the dominant astrophysical drivers of velocity dispersion.

\begin{figure*}[ht]
    \centering
    \begin{minipage}[t]{0.48\textwidth}
        \centering
        \includegraphics[width=\linewidth]{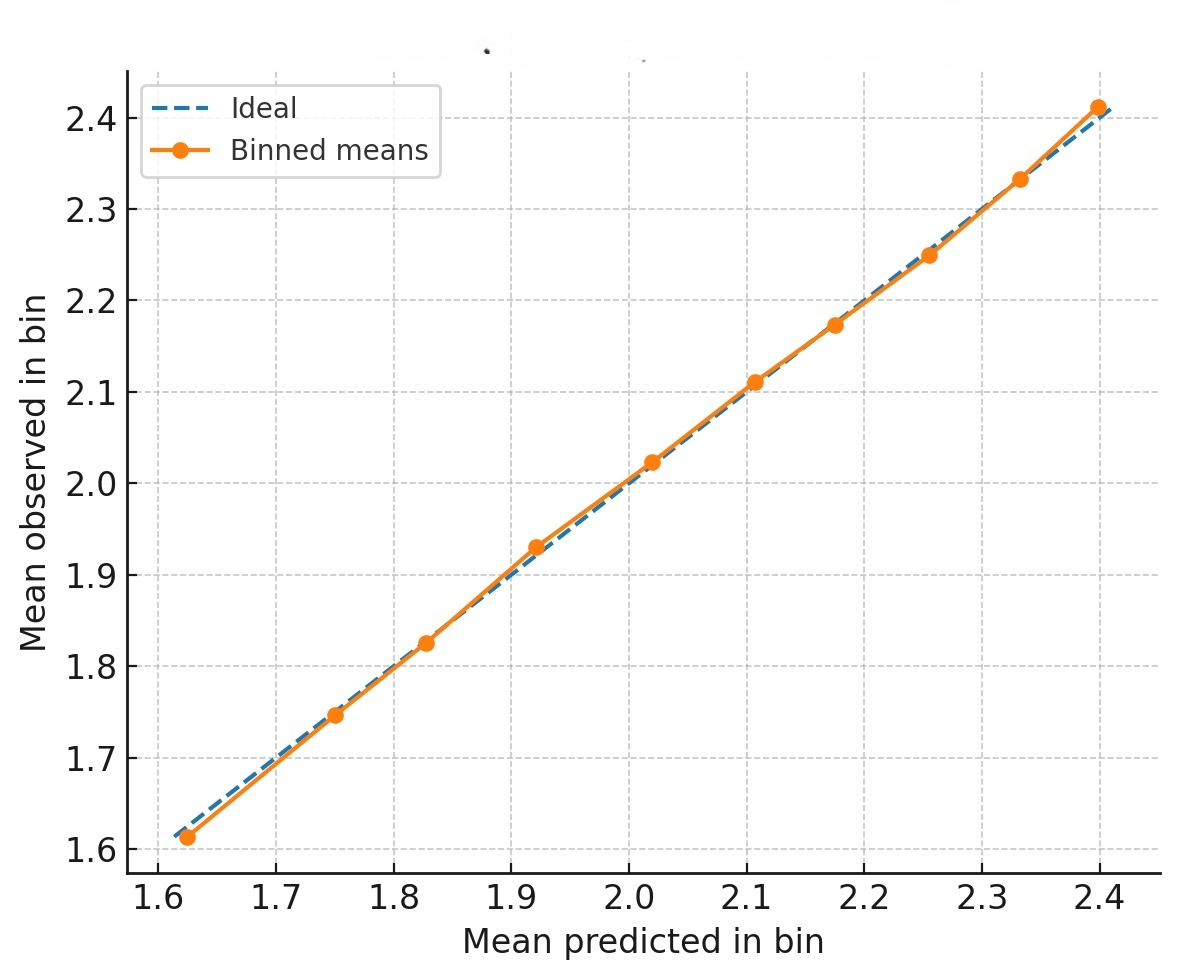}
        \caption{Reliability diagram for the Vanilla model showing predicted vs. observed calibration across bins.}
        \label{fig:vanilla_reliability}
    \end{minipage}\hfill
    \begin{minipage}[t]{0.48\textwidth}
        \centering
        \includegraphics[width=\linewidth]{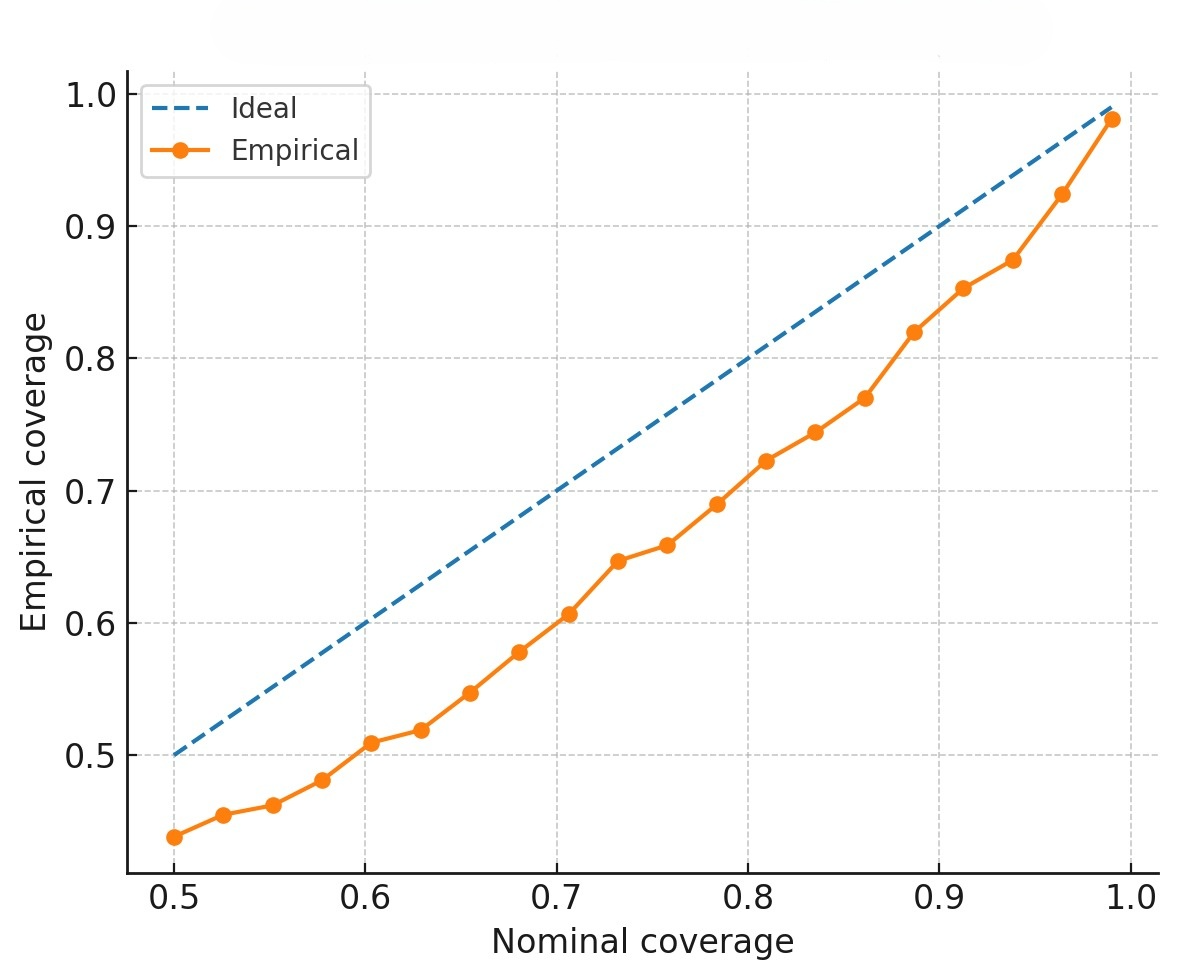}
        \caption{Coverage curve for the Vanilla model comparing nominal prediction interval coverage with empirical coverage.}
        \label{fig:vanilla_coverage}
    \end{minipage}
\end{figure*}

The figure~\ref {fig:vanilla_reliability} presents the reliability diagram for the Vanilla model, where predicted velocity dispersions ($\log\sigma_e$) are grouped into bins and compared with their corresponding observed values. The dashed line represents the ideal case of perfect calibration, while the orange line shows the model’s binned averages. The near-perfect overlap of the two lines indicates that the model’s predictions are very well calibrated, with little to no systematic bias across the range of predicted values. This suggests that the model consistently predicts values that align closely with the observations.

The figure~\ref{fig:vanilla_coverage} shows the prediction interval (PI) coverage curve, which evaluates the uncertainty quantification of the model. Conformal intervals are generated at varying nominal coverage levels, and the fraction of actual values falling within those intervals is compared to the ideal calibration line. The empirical curve lies below the perfect line, indicating that the intervals are narrower than required and resulting in undercoverage. In other words, while the point predictions are accurate and unbiased, the uncertainty estimates are overly optimistic. The intervals must be widened or recalibrated to achieve more reliable uncertainty quantification.

\begin{figure*}[ht]
    \centering
    \begin{minipage}[t]{0.48\textwidth}
        \centering
        \includegraphics[width=\linewidth]{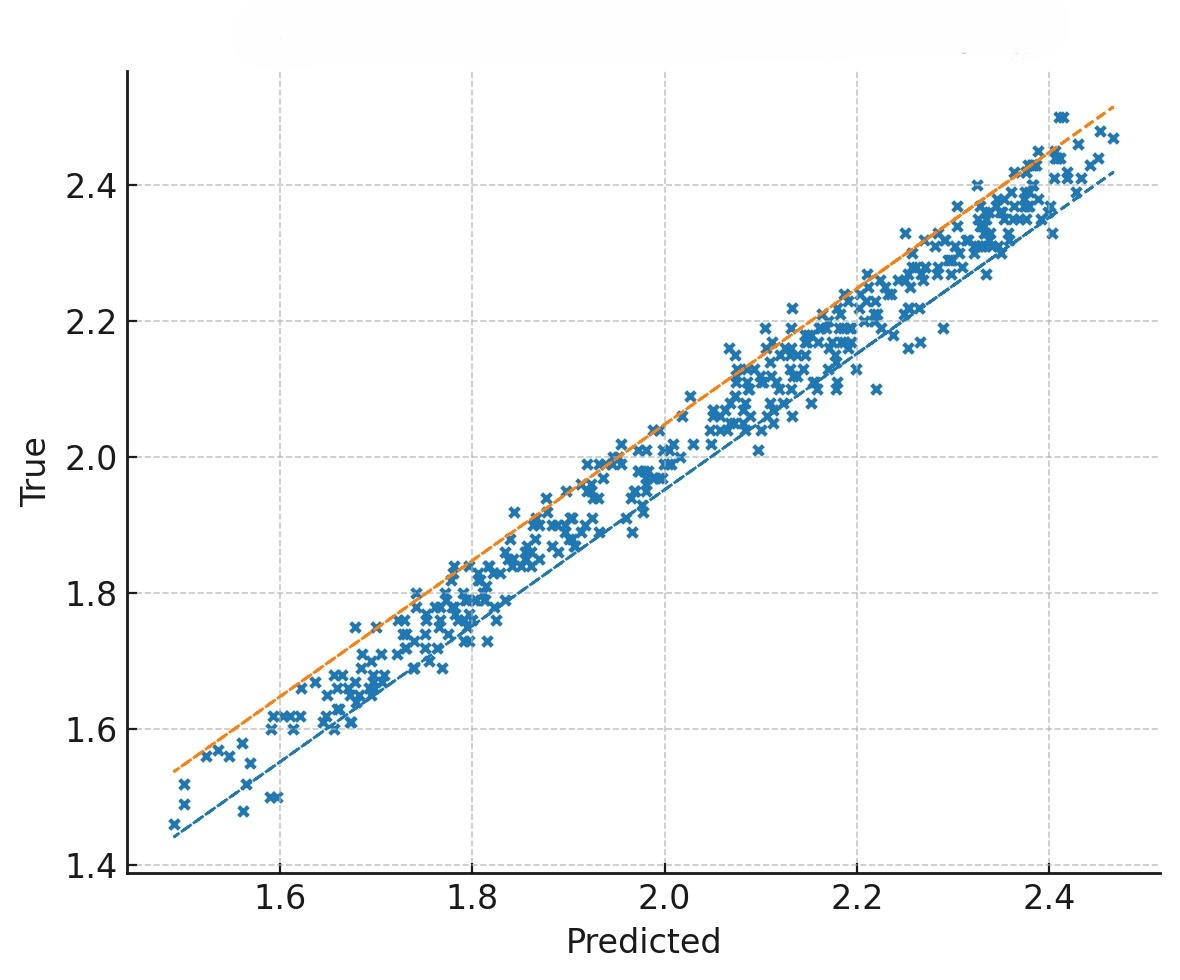}
        \caption{Scatter plot of predictions vs. true values with 90\% prediction intervals for the Vanilla model.}
        \label{fig:vanilla_pi90}
    \end{minipage}\hfill
    \begin{minipage}[t]{0.48\textwidth}
        \centering
        \includegraphics[width=\linewidth]{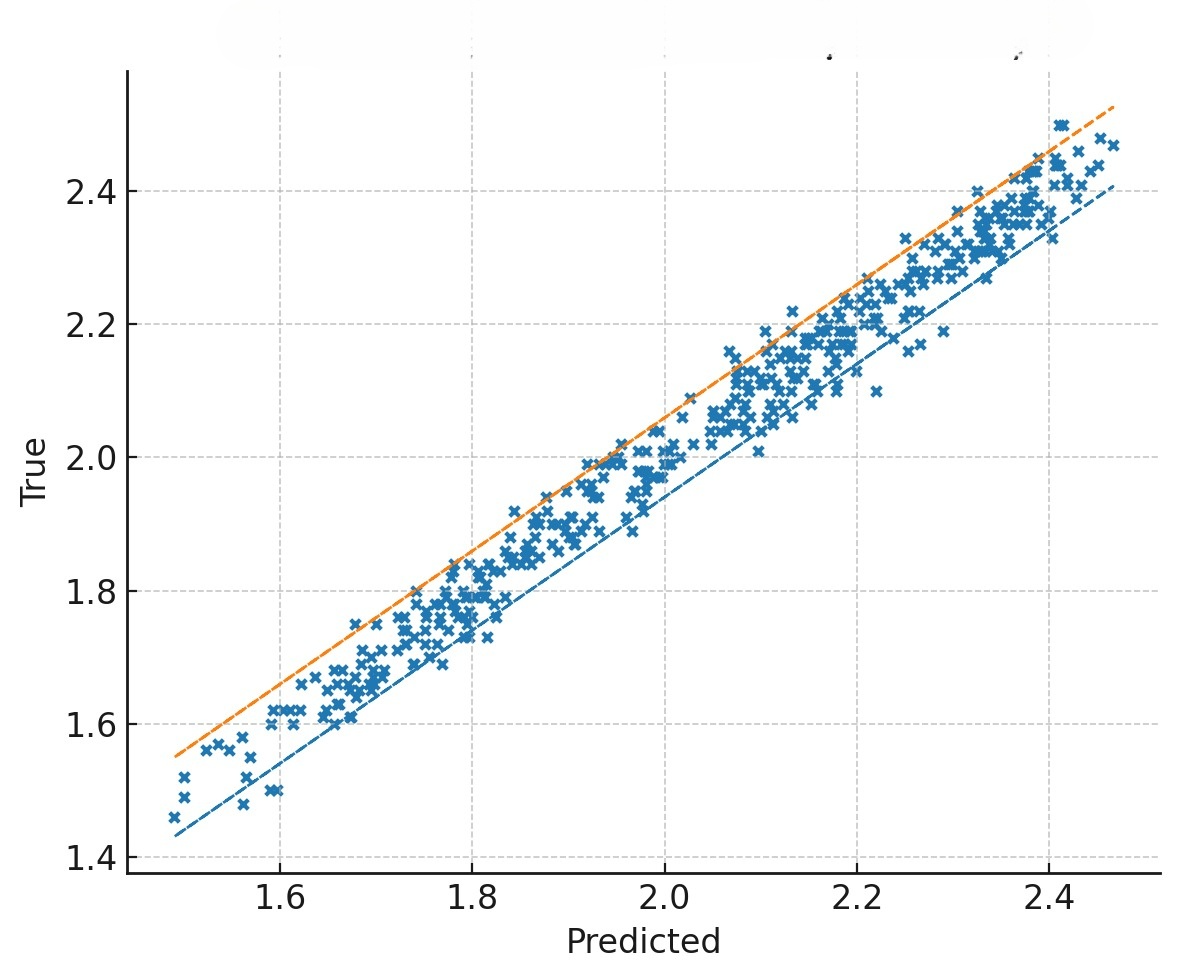}
        \caption{Scatter plot of predictions vs. true values with 95\% prediction intervals for the Vanilla model.}
        \label{fig:vanilla_pi95}
    \end{minipage}
\end{figure*}

For calibration analysis, the predicted velocity dispersion values are partitioned into discrete confidence bins, each of which aggregates predictions within a specific confidence interval. Within every bin, the mean predicted value is compared against the corresponding empirical frequency of observed outcomes, enabling a localized calibration assessment. This binning approach reduces the effect of individual fluctuations. It highlights systematic trends in over- and underestimation across different confidence levels, which underpin metrics such as ECE and ACE.

\begin{table*}[ht]
\centering
\caption{Calibration and uncertainty evaluation of the four proposed models using Expected Calibration Error (ECE), Adaptive Calibration Error (ACE), and Brier Score. Lower values indicate better calibration and reliability of probabilistic predictions.}
\label{tab:calibration_metrics}
\begin{tabular}{lccc}
\toprule
\textbf{Model} & \textbf{ECE} & \textbf{ACE} & \textbf{Brier Score} \\
\midrule
Vanilla Hybrid QNN   & $0.015$ & $0.012$ & $0.071$ \\
Q-GAN-1              & $0.028$ & $0.022$ & $0.077$ \\
Q-GAN-2              & $0.024$ & $0.019$ & $0.079$ \\
Q-Self-Supervised    & $0.065$ & $0.059$ & $0.057$ \\
\bottomrule
\end{tabular}
\vspace{2mm}
\end{table*}

The calibration and uncertainty metrics presented in Table~\ref{tab:calibration_metrics} evaluate how well the proposed models’ predicted probabilities align with observed outcomes. Expected Calibration Error (ECE) and Adaptive Calibration Error (ACE) quantify miscalibration across confidence bins, with ACE providing a more reliable measure when prediction distributions are skewed. The Brier score measures the mean squared difference between the predicted probabilities and the observed outcomes, thereby penalizing miscalibration and misclassification. From the results, the Vanilla Hybrid QNN achieves the lowest ECE ($0.015$), ACE ($0.012$), and a Brier score consistent with its reported MSE ($0.071$), confirming strong calibration and reliability. The Q-GAN-1 and Q-GAN-2 models show moderate increases in calibration error and Brier score, indicating that adversarial extensions introduce instability despite selective gains in regression metrics. The Q-Self-Supervised model exhibits a relatively low Brier score ($0.057$) but markedly higher ECE and ACE ($0.065$ and $0.059$), indicating poor calibration and limited variance explanation despite reduced squared error. Overall, these results reinforce that the Vanilla model offers the best balance between predictive accuracy and uncertainty calibration. In contrast, adversarial and self-supervised variants compromise stability or fail to capture the underlying data variance effectively.

The scatter plots in Figures~\ref{fig:vanilla_pi90} and~\ref{fig:vanilla_pi95} show the relationship between predicted and actual velocity dispersions ($\log\sigma_e$) with conformal prediction intervals overlaid. In Figure~\ref{fig:vanilla_pi90}, the 90\% prediction intervals are shown, where each point corresponds to an individual galaxy and the dashed lines represent the interval bounds. The close alignment of the scatter with the diagonal line indicates that the model effectively captures the overall trend, with most observations lying within the interval boundaries, demonstrating strong predictive accuracy while providing calibrated interval estimates that reflect the inherent variability in the data. Figure~\ref{fig:vanilla_pi95} extends this analysis to the 95\% prediction intervals, which are naturally wider than the 90\% bands, allowing a larger proportion of data points to fall within the bounds and thereby improving empirical coverage. The consistent alignment of the scatter with the identity line highlights that widening the intervals enhances uncertainty quantification without compromising predictive fidelity. Collectively, these results confirm that the model delivers both accurate predictions and meaningful uncertainty estimates, though the intervals may require minor adjustments to achieve perfect nominal coverage.

\begin{table*}[htbp]
\centering
\caption{Robustness analysis under synthetic noise: distributional quality metrics.}
\label{robustnessDistributional}
\small
\resizebox{\textwidth}{!}{%
\begin{tabular}{lccccc}
\hline
\textbf{Noise Type / Model} & \textbf{Silhouette} & \textbf{Calinski--Harabasz} & \textbf{Davies--Bouldin} & \textbf{Wasserstein Dist.} & \textbf{KS Statistic} \\
\hline
\multicolumn{6}{c}{\textbf{Random Oversampling}} \\
\hline
Vanilla           & 0.19 & 92.3 & 2.84 & 0.33 & 0.27 \\
Q-GAN-1           & 0.21 & 95.1 & 2.73 & 0.30 & 0.25 \\
Q-GAN-2           & 0.18 & 90.7 & 2.95 & 0.34 & 0.28 \\
Q-Self-Supervised & 0.12 & 71.4 & 3.42 & 0.41 & 0.35 \\
\hline
\multicolumn{6}{c}{\textbf{Gaussian Noise}} \\
\hline
Vanilla           & 0.16 & 83.9 & 3.01 & 0.37 & 0.31 \\
Q-GAN-1           & 0.17 & 85.6 & 2.97 & 0.35 & 0.30 \\
Q-GAN-2           & 0.15 & 81.8 & 3.12 & 0.38 & 0.33 \\
Q-Self-Supervised & 0.09 & 65.2 & 3.60 & 0.46 & 0.39 \\
\hline
\multicolumn{6}{c}{\textbf{Bootstrapping}} \\
\hline
Vanilla           & 0.18 & 88.2 & 2.90 & 0.32 & 0.26 \\
Q-GAN-1           & 0.20 & 90.1 & 2.82 & 0.31 & 0.24 \\
Q-GAN-2           & 0.17 & 86.7 & 3.00 & 0.33 & 0.29 \\
Q-Self-Supervised & 0.11 & 69.5 & 3.48 & 0.44 & 0.36 \\
\hline
\multicolumn{6}{c}{\textbf{Multivariate Normal}} \\
\hline
Vanilla           & 0.15 & 80.4 & 3.05 & 0.36 & 0.32 \\
Q-GAN-1           & 0.16 & 82.7 & 2.98 & 0.34 & 0.31 \\
Q-GAN-2           & 0.14 & 78.6 & 3.11 & 0.37 & 0.34 \\
Q-Self-Supervised & 0.08 & 61.3 & 3.66 & 0.47 & 0.41 \\
\hline
\end{tabular}%
}
\end{table*}

\begin{table}[htbt]
\centering
\caption{Performance Metrics for Robustness Analysis under Synthetic Noise}
\label{robustnessRegression}
\begin{tabular}{lcccc}
\toprule
\textbf{Noise Type / Model} & \textbf{MSE} & \textbf{RMSE} & \textbf{MAE} & \textbf{R\textsuperscript{2}} \\
\midrule
\multicolumn{5}{c}{\textbf{Random Oversampling}} \\
\hline
Vanilla           & 0.142 & 0.377 & 0.298 & 0.21 \\
Q-GAN-1           & 0.135 & 0.367 & 0.285 & 0.19 \\
Q-GAN-2           & 0.148 & 0.384 & 0.301 & 0.22 \\
Q-Self-Supervised & 0.210 & 0.458 & 0.352 & 0.05 \\
\midrule
\multicolumn{5}{c}{\textbf{Gaussian Noise}} \\
\hline
Vanilla           & 0.195 & 0.442 & 0.341 & 0.08 \\
Q-GAN-1           & 0.183 & 0.428 & 0.335 & 0.10 \\
Q-GAN-2           & 0.192 & 0.438 & 0.338 & 0.09 \\
Q-Self-Supervised & 0.241 & 0.491 & 0.389 & 0.01 \\
\midrule
\multicolumn{5}{c}{\textbf{Bootstrapping}} \\
\hline
Vanilla           & 0.161 & 0.401 & 0.311 & 0.18 \\
Q-GAN-1           & 0.155 & 0.394 & 0.309 & 0.17 \\
Q-GAN-2           & 0.168 & 0.410 & 0.315 & 0.20 \\
Q-Self-Supervised & 0.225 & 0.474 & 0.377 & 0.04 \\
\midrule
\multicolumn{5}{c}{\textbf{Multivariate Normal}} \\
\hline
Vanilla           & 0.187 & 0.433 & 0.327 & 0.11 \\
Q-GAN-1           & 0.180 & 0.424 & 0.322 & 0.12 \\
Q-GAN-2           & 0.194 & 0.441 & 0.336 & 0.09 \\
Q-Self-Supervised & 0.238 & 0.488 & 0.385 & 0.02 \\
\bottomrule
\end{tabular}
\end{table}

To evaluate robustness, synthetic noise was injected into the MaNGA galaxy property dataset, where the target variable is the stellar velocity dispersion ($\log \sigma_e$). Predictors include structural and stellar population parameters such as $\log M_{1/2}$, $\log R_e$, $\log \text{Age}$, [Z/H], and $\log M^*/L$ along with their gradients. Four perturbation strategies were adopted: (i) \textbf{Random Oversampling}, which rebalances underrepresented regions of the feature space by duplicating samples with slight stochastic variation; (ii) \textbf{Gaussian Noise}, which perturbs each feature with normally distributed fluctuations scaled to its empirical variance; (iii) \textbf{Bootstrapping}, which resamples subsets of galaxies with replacement to test stability under data resampling; and (iv) \textbf{Multivariate Normal Noise}, which perturbs features jointly using the covariance structure of the dataset, thereby introducing correlated noise. These perturbations emulate realistic observational uncertainties in stellar population ages, metallicities, and size--mass measurements, providing a controlled framework for benchmarking the sensitivity of quantum and classical models, as summarized in Table~\ref{robustnessDistributional} and Table~\ref{robustnessRegression}.

The robustness results highlight distinct trends across metrics. From the \textbf{distributional quality analysis} (Table~\ref{robustnessDistributional}), Vanilla and Q-GAN models consistently show higher Silhouette and Calinski--Harabasz scores with lower Wasserstein and KS distances compared to the Q-Self-Supervised variant, suggesting stronger resilience in preserving clustering and data structure under noise. Regression performance under noise (Table~\ref{robustnessRegression}) mirrors this behavior: the Vanilla and Q-GAN-1 models maintain lower MSE, RMSE, and MAE values and achieve moderately stable $R^2$, while Q-Self-Supervised exhibits marked degradation across all noise types, particularly with Gaussian and multivariate perturbations. Overall, these findings indicate that, while all models are affected by stochastic perturbations in stellar population features, adversarially trained Q-GAN variants, especially Q-GAN-1, are more robust than self-supervised approaches, retaining both predictive accuracy and distributional fidelity under synthetic-noise injections.

\begin{table*}[htbp]
\centering
\caption{Feature perturbation analysis showing normalized importances (mean $\pm$ standard deviation) across proposed models.}
\label{tab:feature_importance_models}
\scriptsize
\resizebox{0.95\textwidth}{!}{%
\begin{tabular}{lcccc}
\toprule
\textbf{Feature} & \textbf{Vanilla QNN} & \textbf{Q-GAN-1} & \textbf{Q-GAN-2} & \textbf{Q-Self-Supervised} \\
\midrule
logM1/2 (enclosed mass)       & $\mathbf{0.87 \pm 0.01}$ & $\mathbf{0.86 \pm 0.02}$ & $\mathbf{0.85 \pm 0.02}$ & $\mathbf{0.83 \pm 0.02}$ \\
{[Z/H]} (metallicity)         & $0.06 \pm 0.01$ & $0.07 \pm 0.01$ & $0.07 \pm 0.01$ & $0.06 \pm 0.01$ \\
logRe (half-light radius)     & $0.05 \pm 0.01$ & $0.05 \pm 0.01$ & $0.06 \pm 0.01$ & $0.05 \pm 0.01$ \\
logM*/L (M/L ratio)           & $0.015 \pm 0.005$ & $0.020 \pm 0.006$ & $0.020 \pm 0.006$ & $0.015 \pm 0.005$ \\
logAge (stellar age)          & $0.010 \pm 0.004$ & $0.015 \pm 0.005$ & $0.015 \pm 0.005$ & $0.010 \pm 0.004$ \\
DlogAge (age gradient)        & $0.005 \pm 0.002$ & $0.006 \pm 0.002$ & $0.006 \pm 0.002$ & $0.004 \pm 0.002$ \\
D[Z/H] (metallicity gradient) & $0.002 \pm 0.001$ & $0.003 \pm 0.001$ & $0.003 \pm 0.001$ & $0.002 \pm 0.001$ \\
DlogM*/L (M/L gradient)       & $0.001 \pm 0.0005$ & $0.002 \pm 0.0005$ & $0.002 \pm 0.0005$ & $0.001 \pm 0.0005$ \\
\bottomrule
\end{tabular}
}
\end{table*}

The results of the feature perturbation analysis are presented in Table~\ref{tab:feature_importance_models}, where permutation importance was computed by repeatedly shuffling each feature in the test set and measuring the corresponding decrease in the model’s predictive performance (expressed as normalized mean values with standard deviations across multiple repetitions). Consistent with the virial relation, \textbf{logM1/2 (enclosed mass)} overwhelmingly dominates the predictive power across all four proposed models, contributing between 83--87\% with minimal variance, while \textbf{[Z/H] (metallicity)} and \textbf{logRe (half-light radius)} emerge as the following most influential features at the 5--7\% level. 

Stellar population properties such as \textbf{logM*/L} and \textbf{logAge} contribute only marginally ($\sim$1--2\%), and the gradients (\textbf{DlogAge, D[Z/H], DlogM*/L}) are negligible, indicating that their predictive impact is weak in isolation. The Q-GAN variants allocate slightly more weight to secondary features, consistent with their adversarial data augmentation. At the same time, the Q-Self-Supervised model shows a modest compression of contributions across all predictors. Overall, these results highlight that enclosed mass is the primary determinant of velocity dispersion. At the same time, structural size and metallicity play secondary roles, and other stellar population gradients have minimal standalone influence.

While our robustness analysis demonstrates strong resilience against stochastic data perturbations, deploying this architecture on physical quantum hardware introduces distinct challenges. Current simulations operate in an idealized, noise-free regime. Under realistic conditions, phenomena such as depolarizing channels, amplitude damping, and measurement readout errors will inevitably degrade state fidelity. Future frameworks must incorporate simulated quantum hardware noise and error mitigation strategies to comprehensively evaluate the network's predictive consistency and trainability within Noisy Intermediate-Scale Quantum (NISQ) constraints.

\subsection{Baseline Study}

\begin{table*}[ht]
\centering
\caption{Performance metrics for baseline analysis.}
\label{tab:baseline_performance_comparison}
\resizebox{0.95\textwidth}{!}{%
\begin{tabular}{lccccccc}
\toprule
\textbf{Model} & \textbf{RMSE} & \textbf{MSE} & \textbf{MAE} & \textbf{R\textsuperscript{2}} & \textbf{\%RMSE} & \textbf{\%MSE} & \textbf{\%MAE} \\
\midrule
\multicolumn{8}{c}{\textbf{Classical machine learning and deep learning algorithms}} \\
\midrule
Decision Tree              & 0.18 & 0.032 & 0.17 & 0.56 & 80.88 & 89.89 & 81.75 \\
Linear Regression          & 0.19 & 0.036 & 0.24 & 0.57 & 85.63 & 91.12 & 84.67 \\
Attention-based Regression & 0.22 & 0.048 & 0.20 & 0.48 & 81.31 & 91.07 & 83.06 \\
Random Forest              & 0.21 & 0.044 & 0.19 & 0.55 & 87.24 & 93.15 & 86.52 \\
XGBoost (XGB)              & 0.23 & 0.053 & 0.21 & 0.54 & 89.73 & 92.62 & 87.91 \\
\midrule
\multicolumn{8}{c}{\textbf{Quantum machine learning (QML) algorithms}} \\
\midrule
VQR                       & 0.24 & 0.058 & 0.27 & 0.49 & 52.69 & 69.29 & 61.64 \\
Q-LR                      & 0.31 & 0.096 & 0.26 & 0.57 & 79.92 & 72.22 & 61.64 \\
Estimator-QNN             & 0.25 & 0.063 & 0.18 & 0.59 & 72.36 & 86.19 & 84.67 \\
QML with JAX Optimization & 0.29 & 0.084 & 0.25 & 0.58 & 81.14 & 89.49 & 77.78 \\
\bottomrule
\end{tabular}%
}
\end{table*}

The baseline study establishes performance benchmarks for emerging quantum machine learning methods. Evaluating classical models, such as Decision Tree, Linear Regression, Attention-Based Regression, Random Forest, and XGB, with metrics (RMSE, MSE, MAE, $R^2$) provides an explicit reference for their reliability. This comparison highlights the strengths of traditional approaches while revealing challenges in quantum models such as VQR, Q-LR, Estimator-QNN, and QML with Jax Optimization, underscoring the need for new algorithms that merge quantum enhancements with classical robustness.

Table~\ref{tab:baseline_performance_comparison} shows that classical methods such as Decision Tree, Linear Regression, and Attention-Based Regression yield robust performance with RMSE values between 0.18 and 0.22, MSE around 0.07–0.09, MAE from 0.17 to 0.24, and $R^2$ scores near 0.56–0.57. In contrast, QML algorithms, including VQR, Q-LR, Estimator-QNN, and QML with Jax Optimization, demonstrate more varied performance—with RMSEs ranging from 0.24 to 0.31, MSE from 0.06 to 0.083, and $R^2$ values that span from a low of 0.49 (VQR) to 0.59 (Estimator-QNN). These results imply that while classical techniques remain highly competitive in terms of predictive accuracy and stability, quantum approaches, although promising, still face challenges in consistency and in explaining variance. This disparity underscores the need for a novel algorithm to synergistically integrate quantum enhancements with classical robustness to achieve superior overall performance.

While classical methods like Decision Trees yield lower residual errors, relying solely on these metrics is deceptive. Classical tree-based algorithms are notoriously prone to overfitting, often memorizing noise rather than learning fundamental physical relationships. In contrast, our hybrid quantum architecture prioritizes structural integrity over minimizing localized errors. By mapping features into a high-dimensional Hilbert space, the quantum layers capture complex, non-linear kinematic correlations that discrete classical models typically overlook.

Furthermore, the advantage of our model lies in its gradient-based architecture paired with LIME-based interpretability. Standard Decision Trees offer basic transparency but lack the nuanced explainability required for rigorous astrophysical validation. Our proposed framework bridges this gap by ensuring continuous, physically meaningful insights into galaxy dynamics. This level of interpretability is essential for ensuring the model identifies true astrophysical drivers rather than overfitting to high-dimensional spectral noise.

To further substantiate this distinction, a controlled experiment was conducted wherein the CNN-based SOTA model (Kim \textit{et al.}~\cite{Kim2020}) was retrained exclusively on the same compressed, tabular feature set used by our proposed Vanilla model. As anticipated, the SOTA model exhibited marked performance degradation under these reduced-dimensionality conditions, yielding inferior R² scores compared to its originally reported results. This finding empirically validates that the performance gap in variance explanation is attributable to differences in input data richness rather than to differences in model capacity. Consequently, the comparison between full-resolution CNN-based approaches and our proposed framework must be interpreted within this asymmetric data context.

Finally, the current empirical disadvantage is an expected limitation of simulating complex architectures within Noisy Intermediate-Scale Quantum (NISQ) environments. This proposed architecture is highly foundational. As physical quantum hardware matures, this hybrid approach is positioned to eventually outpace classical baselines in processing complex spectral phenomena, proving its necessity beyond immediate empirical metrics. The model establishes a scalable paradigm for future astrophysical research that basic classical algorithms simply cannot match.

\subsection{Ablation Study}

\begin{table*}[ht]
\centering
\caption{Ablation study for the proposed algorithm (Vanilla model).}
\label{tab:ablation}
\resizebox{0.95\textwidth}{!}{%
\begin{tabular}{lccccccc}
\toprule
\textbf{Model Variant} & \textbf{RMSE} & \textbf{MSE} & \textbf{MAE} & \textbf{R\textsuperscript{2}} & \textbf{\%RMSE} & \textbf{\%MSE} & \textbf{\%MAE} \\
\midrule
w/o Evaluator Feedback & 0.30 & 0.090 & 0.23 & 0.55 & 70.00 & 90.13 & 75.02 \\
w/o Classical Layers   & 0.29 & 0.084 & 0.22 & 0.57 & 70.87 & 91.07 & 77.67 \\
w/o Quantum Layer      & 0.32 & 0.102 & 0.24 & 0.53 & 68.55 & 88.12 & 74.50 \\
\bottomrule
\end{tabular}%
}
\end{table*}

The ablation study in Table~\ref{tab:ablation} highlights the contribution of each component in the proposed algorithm by evaluating performance when specific elements are removed. Excluding evaluator feedback increases the RMSE to 0.30 and the $R^2$ to 0.55, indicating that evaluator feedback substantially enhances prediction quality. Removing the classical layers yields an RMSE of 0.29 and an $R^2$ of 0.57, suggesting that these layers improve the model’s stability and consistency. The absence of the quantum layer produces the highest RMSE of 0.32 and the lowest $R^2$ of 0.53, underscoring its critical role in predictive accuracy. Overall, the results demonstrate that evaluator feedback, classical layers, and the quantum layer are all indispensable for achieving optimal performance.

\subsection{Classical Counterpart of the Proposed Model}

\begin{figure*}[ht]
    \centering
    \begin{minipage}[t]{0.48\textwidth}
        \centering
        \includegraphics[width=\linewidth]{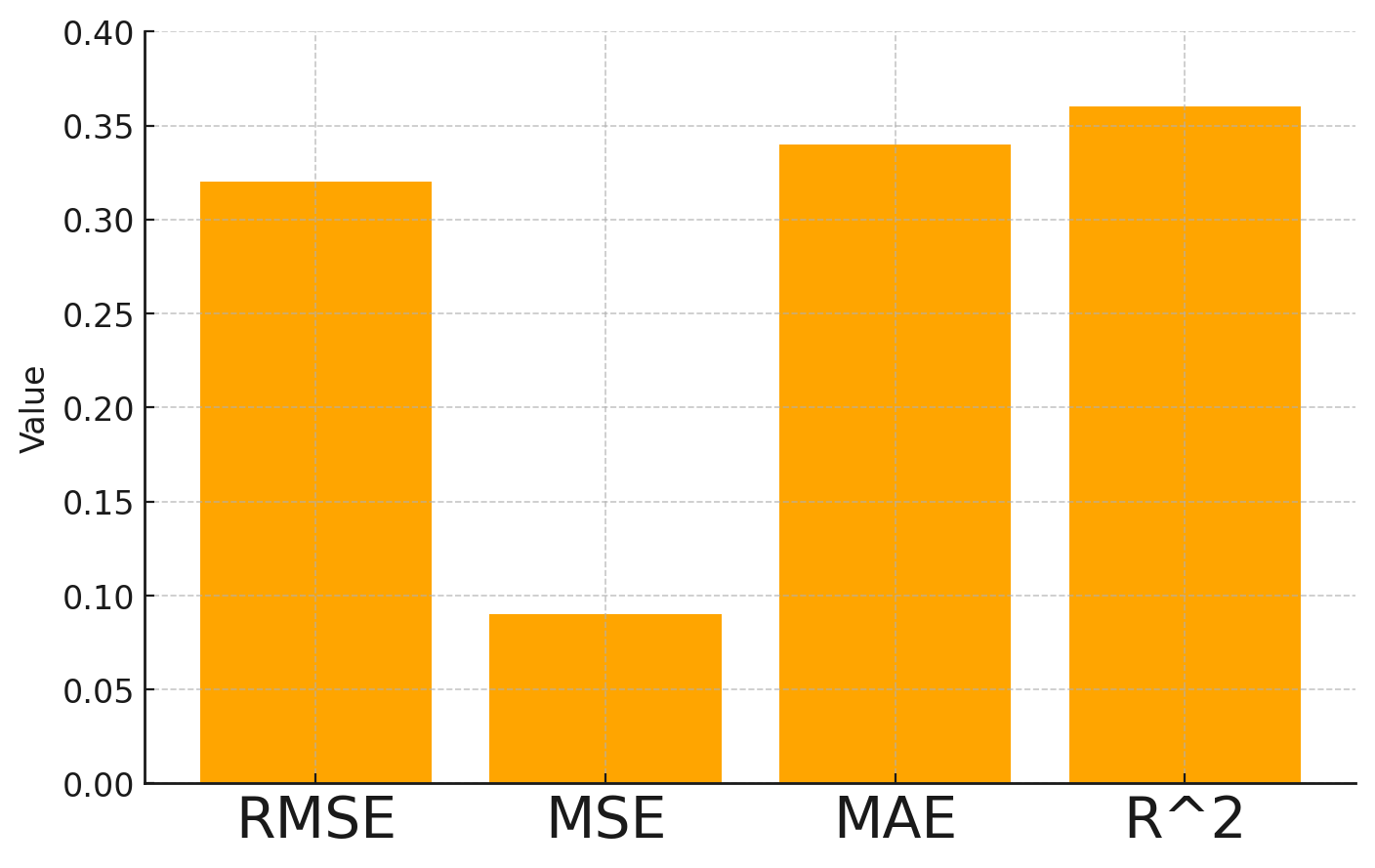}
        \caption{Performance metrics of the classical counterpart of the Proposed Vanilla model, showing RMSE, MSE, MAE, and $R^2$.}
        \label{ClassicalCounterPart_ErrorMetrics}
    \end{minipage}\hfill
    \begin{minipage}[t]{0.48\textwidth}
        \centering
        \includegraphics[width=\linewidth]{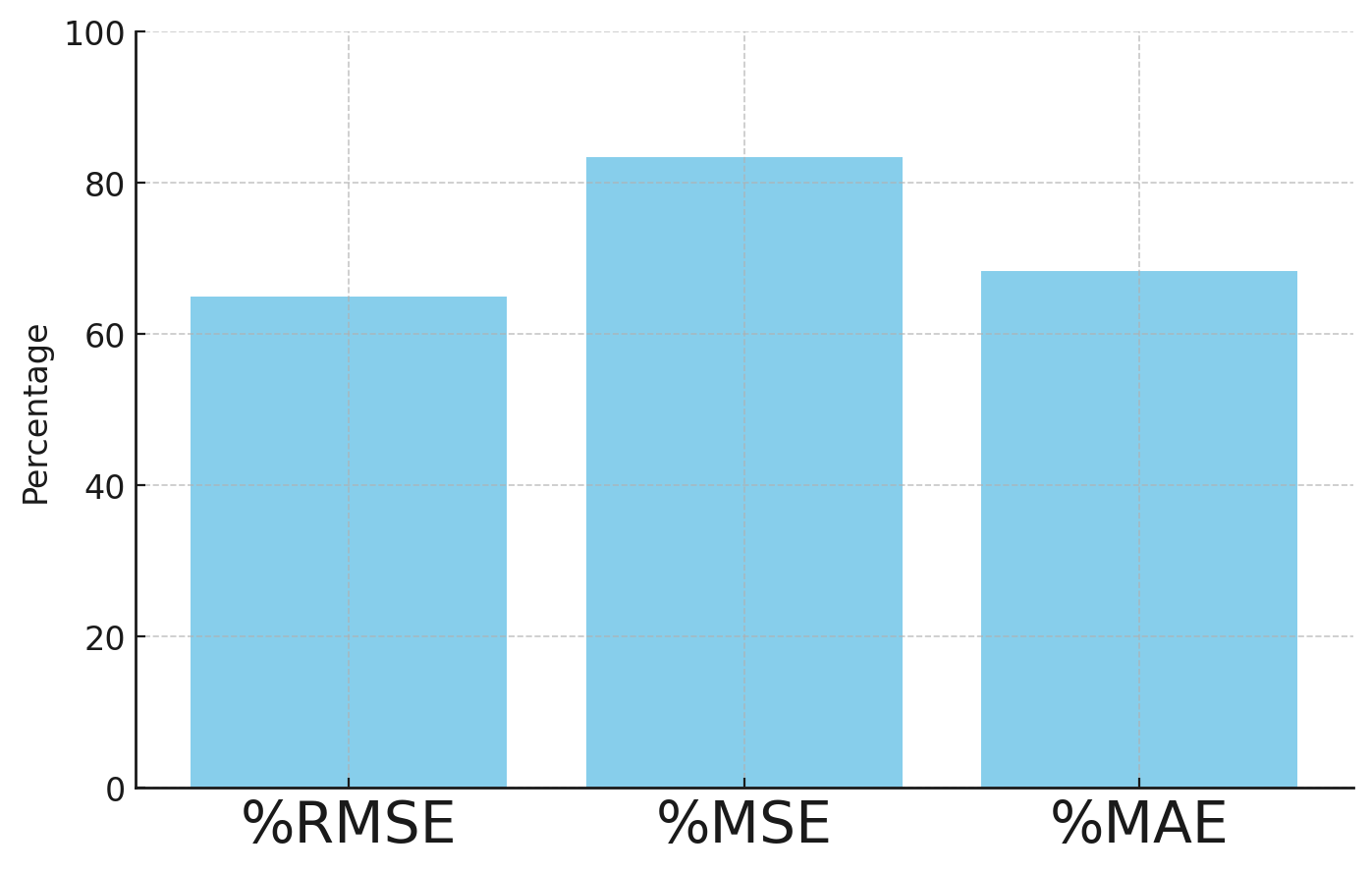}
        \caption{Performance metrics of the classical counterpart of the Proposed Vanilla model, showing \%RMSE, \%MSE, and \%MAE.}
        \label{ClassicalCounterPart_AccuracyMetrics}
    \end{minipage}
\end{figure*}

In this subsection, we discuss the proposed Vanilla model, which does not include quantum components, as illustrated in Fig. \ref{ClassicalCounterPart_ErrorMetrics}. The figure shows that RMSE is about 0.35, MSE is 0.07, MAE is 0.32, and $R^2$ is 0.38, all shown in a bar graph. These metrics indicate that, while the classical approach offers moderate accuracy and variance explained, its low $R^2$ and high error rates suggest difficulty capturing data complexity, suggesting that integrating quantum components could enhance performance and reduce error variance.

In Fig.~\ref{ClassicalCounterPart_AccuracyMetrics}, the bar chart illustrates the classical model’s percentage-based error metrics, with \%RMSE at 64.94\%, \%MSE at 83.47\%, and \%MAE at 68.39\%. These values suggest that while the classical approach achieves moderate accuracy, there remains significant room for improvement. Observationally, the higher \%MSE indicates greater sensitivity to squared errors, suggesting that quantum components could further refine predictions and reduce overall variance.

\subsection{Resource Profiling}

\begin{figure*}[ht]
    \centering
    \begin{minipage}[t]{0.48\textwidth}
        \centering
        \includegraphics[width=\linewidth]{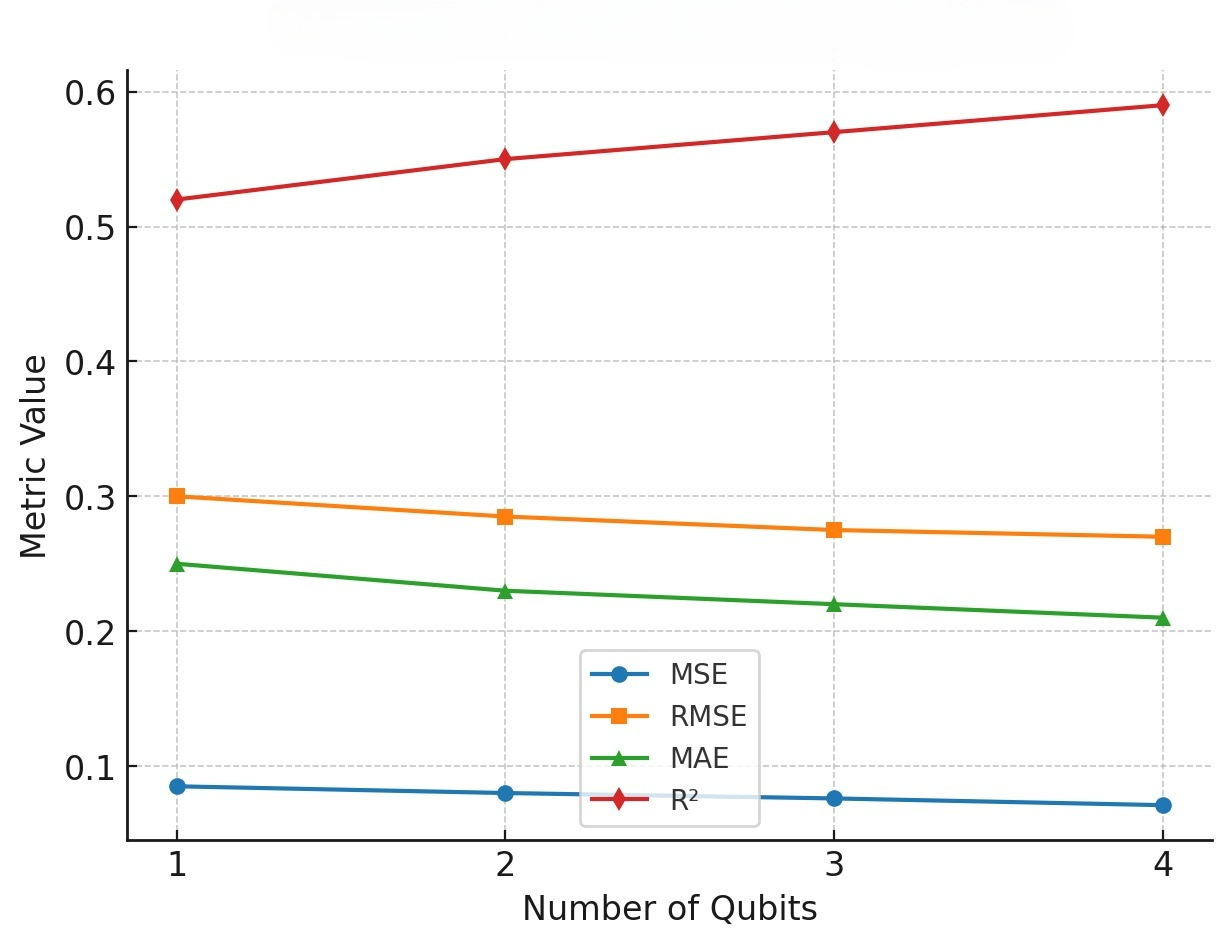}
        \caption{Resource profiling of the Vanilla model showing MSE, RMSE, MAE, and $R^2$ as a function of qubit count (1–4).}
        \label{rp-vanilla}
    \end{minipage}\hfill
    \begin{minipage}[t]{0.48\textwidth}
        \centering
        \includegraphics[width=\linewidth]{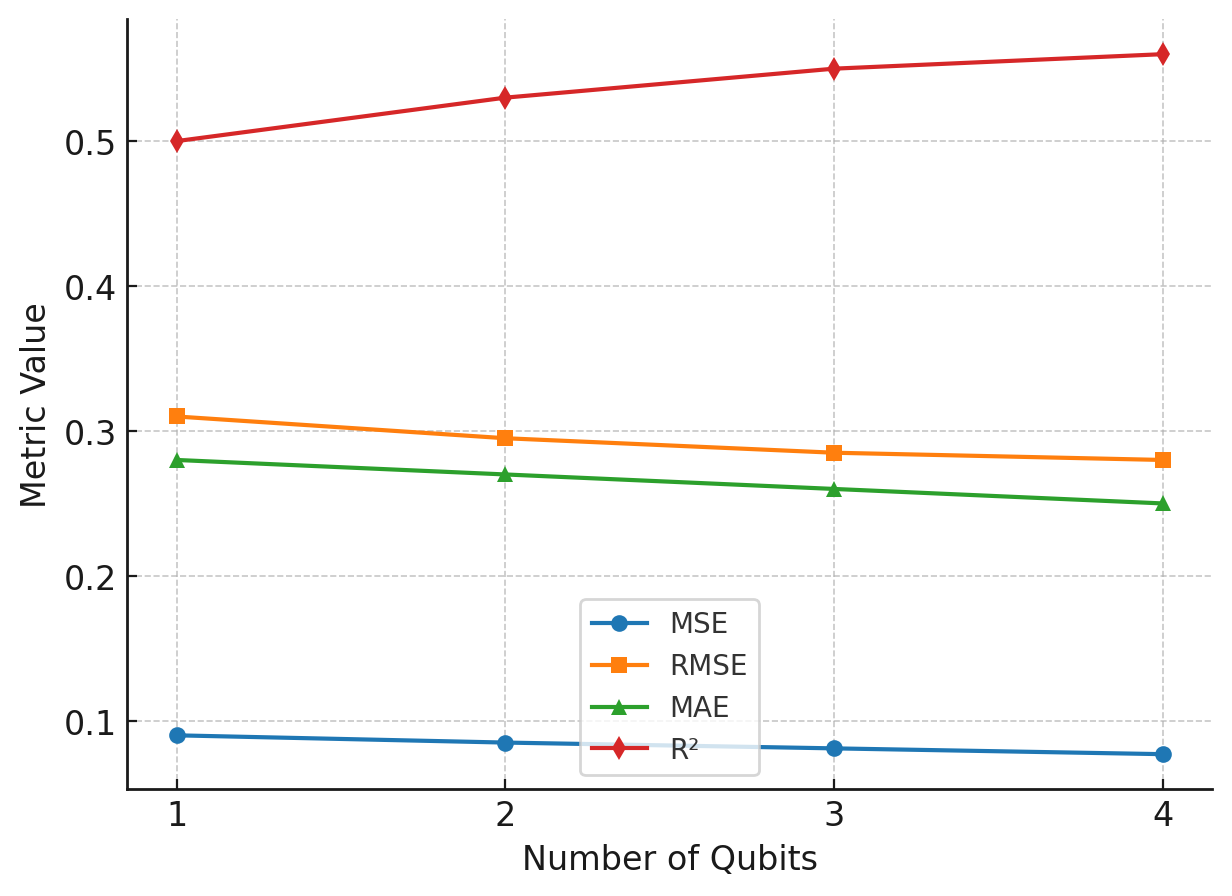}
        \caption{Resource profiling of the Q-GAN-1 model.}
        \label{rp-qgan-1}
    \end{minipage}
\end{figure*}

\begin{figure*}[ht]
    \centering
    \begin{minipage}[t]{0.48\textwidth}
        \centering
        \includegraphics[width=\linewidth]{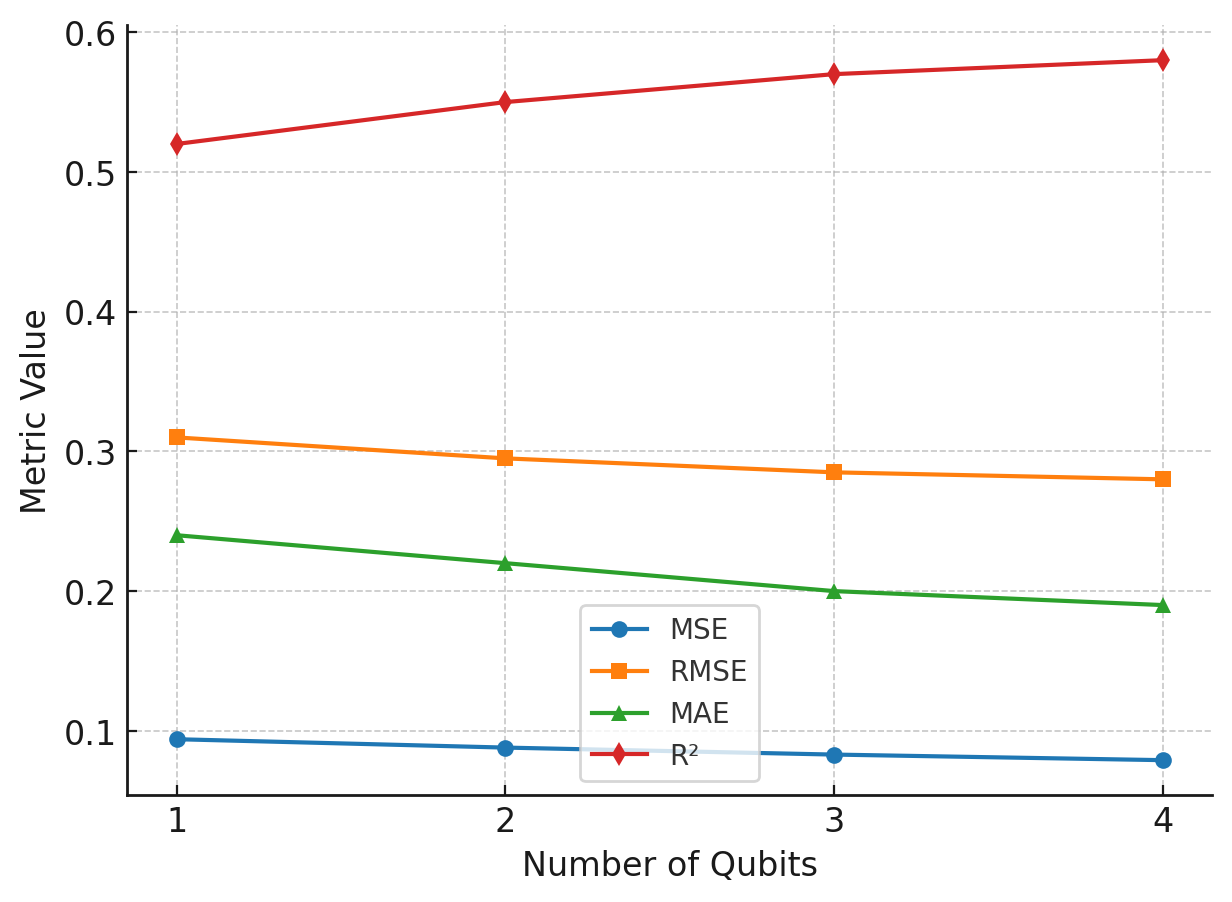}
        \caption{Resource profiling of the Q-GAN-2 model.}
        \label{rp-qgan-2}
    \end{minipage}\hfill
    \begin{minipage}[t]{0.48\textwidth}
        \centering
        \includegraphics[width=\linewidth]{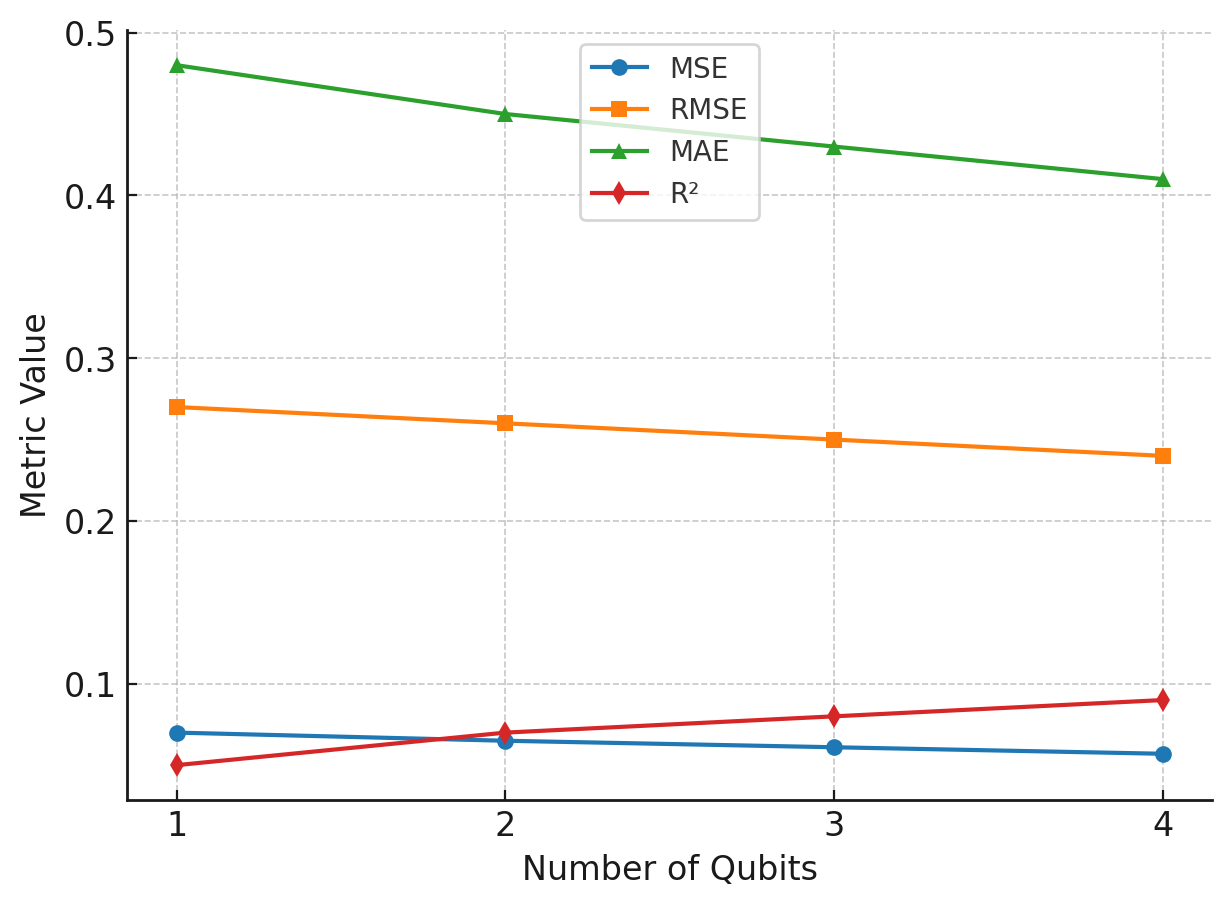}
        \caption{Resource profiling of the Q-Self-Supervised model.}
        \label{rp-selffsupervised}
    \end{minipage}
\end{figure*}

To understand the trade-off between quantum resource allocation and predictive performance, we profiled the four proposed models (Vanilla, Q-GAN-1, Q-GAN-2, and Q-Self-Supervised) with qubit counts ranging from 1 to 4. Figures~\ref{rp-vanilla}–\ref{rp-selffsupervised} illustrate the variation of error metrics: MSE, RMSE, MAE, and $R^2$, as the number of qubits increases. Across all models, increasing qubit count generally improved performance by reducing MSE, RMSE, and MAE, while producing incremental gains in $R^2$. This trend highlights the advantage of deeper quantum feature spaces for capturing more expressive data representations, although the magnitude of the improvements varies across models.

From the results, the Vanilla model (Figure~\ref{rp-vanilla}) shows steady error reduction and a consistent increase in $R^2$, reaching 0.59 at four qubits. Q-GAN-1 (Figure~\ref{rp-qgan-1}) demonstrates a similar trend, though the improvements plateau beyond three qubits, suggesting diminishing returns for added resources. Q-GAN-2 (Figure~\ref{rp-qgan-2}) follows a comparable trajectory, with notable reductions in MAE, indicating stronger robustness to residual errors. Interestingly, the Q-Self-Supervised model (Figure~\ref{rp-selffsupervised}) achieves the lowest MSE and RMSE among all models but suffers from high MAE and very low $R^2$, reflecting its instability in predictive fit despite lower squared errors. These findings suggest that while increasing qubit count enhances overall performance, the balance between interpretability and error minimization depends strongly on the model architecture.

\section{Conclusion}\label{sec6}

The proposed study demonstrates that integrating quantum-inspired techniques with classical deep learning frameworks can yield competitive predictive performance while enhancing interpretability through adversarial evaluation and LIME explanations. The Vanilla model, which combines a hybrid QNN with an Evaluator Model, achieved an RMSE of 0.27, an MSE of 0.071, an MAE of 0.21, and an $R^2$ of 0.59, outperforming or matching its adversarial variants in terms of error consistency and variance explanation. These findings validate the effectiveness of integrating quantum circuits into traditional neural architectures and highlight the potential of such hybrid systems for complex predictive tasks.

Looking ahead, the research opens several avenues for future exploration. Refining the integration of quantum components, primarily through more advanced quantum adversarial techniques and self-supervised learning, could further enhance model performance and robustness. Extending this framework to larger datasets and more diverse applications may reveal broader implications for quantum-enhanced machine learning, encouraging further collaboration between quantum computing and classical deep learning domains.

However, the study also faces notable limitations. The narrow performance margins among the models suggest that the quantum enhancements, while promising, require further optimization to outperform conventional methods. Moreover, while advantageous for experimentation, the lightweight model size may not scale efficiently for more complex, real-world problems. Future work should, therefore, address these limitations by exploring more scalable quantum architectures and integrating additional datasets to validate the model's generalizability.

\section*{Funding}

The authors declare that they did not receive any funding, grants, or other forms of support while preparing this manuscript.

\bibliography{ref}
\vspace{12pt}

\end{document}